\documentclass[journal]{IEEEtran}
\ifCLASSINFOpdf
\else
\fi
\usepackage[cmex10]{amsmath}
\usepackage{array}

\usepackage{mdwmath}
\usepackage{mdwtab}

\usepackage{eqparbox}
\usepackage{fixltx2e}
	\usepackage{url}

\usepackage{subcaption}

\usepackage{float}
\usepackage{color}
\usepackage[noadjust]{cite}
\usepackage{graphicx}
\usepackage{tikz,pgfplots}
\usepackage{graphicx}
\usepackage{epstopdf}
\usepackage{amssymb}
\usepackage{booktabs}
\usepackage{multirow}
\usepackage{multicol}
\usepackage[ruled,vlined]{algorithm2e}
\usepackage{comment}

\begin{document}
%
\title{Multisource and Multitemporal Data Fusion in Remote Sensing}
%
%
%

\author{Pedram~Ghamisi,~\IEEEmembership{Senior Member,~IEEE,}
Behnood~Rasti,~\IEEEmembership{Member,~IEEE,}      
Naoto~Yokoya,~\IEEEmembership{Member,~IEEE,}
Qunming~Wang,
Bernhard~H\"ofle,
Lorenzo~Bruzzone,~\IEEEmembership{Fellow,~IEEE,} 
Francesca~Bovolo,~\IEEEmembership{Senior Member,~IEEE,}
Mingmin~Chi,~\IEEEmembership{Senior Member,~IEEE,}
Katharina~Anders,
Richard~Gloaguen,\\
Peter~M.~Atkinson,
and J\'on~Atli~Benediktsson,~\IEEEmembership{Fellow,~IEEE}
\IEEEcompsocitemizethanks{\IEEEcompsocthanksitem The work of P. Ghamisi is supported by the "High Potential Program" of Helmholtz-Zentrum Dresden-Rossendorf. 

P. Ghamisi and R. Gloaguen are with the Helmholtz-Zentrum Dresden-Rossendorf (HZDR), Helmholtz Institute Freiberg for Resource Technology (HIF), Exploration, D-09599 Freiberg, Germany (emails: p.ghamisi@gmail.com,r.gloaguen@hzdr.de).

B. Rasti is with the Faculty of Electrical and Computer
Engineering, University of Iceland, 107 Reykjavik, Iceland (email: behnood@hi.is).

N. Yokoya is with the RIKEN Center for Advanced Intelligence Project, RIKEN, 103-0027 Tokyo, Japan (e-mail: naoto.yokoya@riken.jp).

Q. Wang is with the College of Surveying and Geo-Informatics, Tongji University, 1239 Siping Road, Shanghai 200092, China (email: wqm11111@126.com).  

B. H\"ofle and K. Anders are with GIScience at the Institute of Geography, Heidelberg University, Germany (emails: hoefle@uni-heidelberg.de, katharina.anders@uni-heidelberg.de).

L. Bruzzone is with the department of Information Engineering and Computer Science, University of Trento, Trento, Italy (email: lorenzo.bruzzone@unitn.it).

F. Bovolo is with the Center for Information and Communication Technology, Fondazione Bruno Kessler, Trento, Italy (email: bovolo@fbk.eu).

M. Chi is with the school of Computer Science, Fudan University, China (email: mmchi@fudan.edu.cn).

P. M. Atkinson is with Lancaster Environment Centre, Lancaster University, Lancaster, U.K (email: pma@lancaster.ac.uk).

J. A. Benediktsson is with the Faculty of Electrical and Computer
Engineering, University of Iceland, 107 Reykjavik, Iceland (e-mail: benedikt@hi.is).

}
\thanks{Manuscript received 2018.}}

%
%

\markboth{IEEE GRSM DRAFT 2018}%
{Shell \MakeLowercase{\textit{et al.}}: Bare Demo of IEEEtran.cls for Journals}
%



\maketitle

\begin{abstract}
\textcolor{blue}{The final version of the paper can be found in IEEE Geoscience and Remote Sensing Magazine.}

The sharp and recent increase in the availability of data captured by different sensors combined with their considerably heterogeneous natures poses a serious challenge for the effective and efficient processing of remotely sensed data. Such an increase in remote sensing and ancillary datasets, however, opens up the possibility of utilizing multimodal datasets in a joint manner to further improve the performance of the processing approaches with respect to the application at hand. Multisource data fusion has, therefore, received enormous attention from researchers worldwide for a wide variety of applications. Moreover, thanks to the revisit capability of several spaceborne sensors, the integration of the temporal information with the spatial and/or spectral/backscattering information of the remotely sensed data is possible and helps to move from a representation of 2D/3D data to 4D data structures, where the time variable adds new information as well as challenges for the information extraction algorithms. There are a huge number of research works dedicated to multisource and multitemporal data fusion, but the methods for the fusion of different modalities have expanded in different paths according to each research community. This paper brings together the advances of multisource and multitemporal data fusion approaches with respect to different research communities and provides a thorough and discipline-specific starting point for researchers at different levels (i.e., students, researchers, and senior researchers) willing to conduct novel investigations on this challenging topic by supplying sufficient detail and references. More specifically, this paper provides a bird's-eye view of many important contributions specifically dedicated to the topics of pansharpening and resolution enhancement, point cloud data fusion, hyperspectral and LiDAR data fusion, multitemporal data fusion, as well as big data and social media. In addition, the main challenges and possible future research for each section are outlined and discussed. 
\end{abstract}


\IEEEpeerreviewmaketitle

\section{Introduction}
\label{Sec:Intro}

The number of data produced by sensing devices has increased exponentially in the last few decades, creating the ``Big Data" phenomenon, and leading to the creation of the new field of ``data science", including the popularization of ``machine learning" and ``deep learning" algorithms to deal with such data \cite{APPEL201847, AUDEBERT201820, Ball2017}. In the field of remote sensing, the number of platforms for producing remotely sensed data has similarly increased, with an ever-growing number of satellites in orbit and planned for launch, and new platforms for proximate sensing such as unmanned aerial vehicles (UAVs) producing very fine spatial resolution data. While optical sensing capabilities have increased in quality and volume, the number of alternative modes of measurement has also grown including, most notably, airborne light detection and ranging (LiDAR) and terrestrial laser scanning (TLS), which produce point clouds representing elevation, as opposed to images \cite{Xia2018LH}. The number of synthetic aperture radar (SAR) sensors, which measure RADAR backscatter, and satellite and airborne hyperspectral sensors, which extend optical sensing capabilities by measuring in a larger number of wavebands, has also increased greatly \cite{EslamiMohammadzadeh2016, Zhu2018}. Airborne and spaceborne geophysical measurements
such as the satellite mission Gravity Recovery And Climate
Experiment (GRACE) or airborne electro-magnetic surveys
are currently been also considered. In addition, there has been great interest in new sources of ancillary data, for example, from social media, crowd sourcing, scraping the internet and so on (\cite{ESTES201641, Foulser2016, Li2017SM}). These data have a very different modality to remote sensing data, but may be related to the subject of interest and, therefore, may add useful information relevant to specific problems.

The remote sensors onboard the above platforms may vary greatly in multiple dimensions; for example, the types of properties sensed and the spatial and spectral resolutions of the data. This is true, even for sensors that are housed on the same platform (e.g., the many examples of multispectral and panchromatic sensors) or that are part of the same satellite configuration (e.g., the European Space Agency's (ESA's) series of Medium Resolution Imaging Spectrometer (MERIS) sensors). The rapid increase in the number and availability of data combined with their deeply heterogeneous natures creates serious challenges for their effective and efficient processing (\cite{Yokoya2018}). For a particular remote sensing application, there are likely to be multiple remote sensing and ancillary datasets pertaining to the problem and this creates a dilemma; how best to combine the datasets for maximum utility? It is for this reason that multisource data fusion, in the context of remote sensing, has received so much attention in recent years \cite{GHASSEMIAN201675, LI2017100, LIU2018158, Yokoya2018}.

\begin{figure*}
\centering
\includegraphics[width=0.999\linewidth]{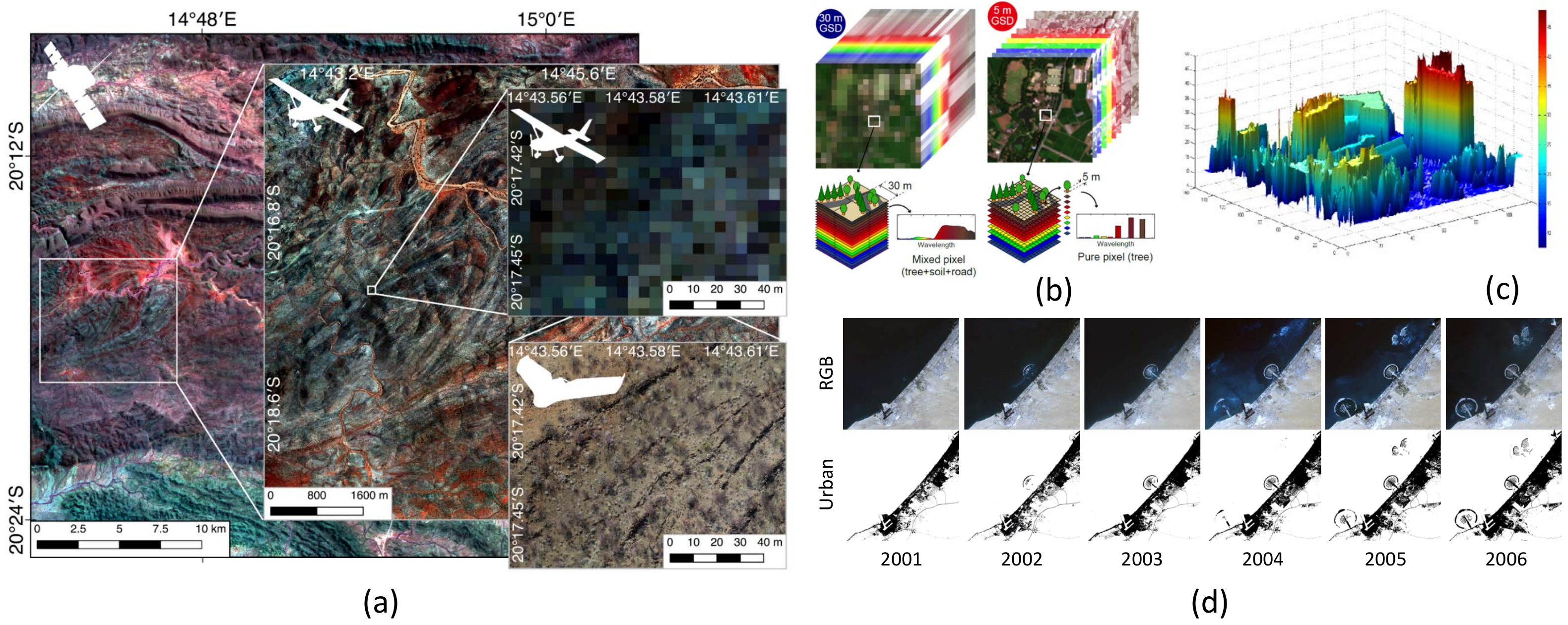}
\caption{(a) The multiscale nature of diverse datasets captured by multisensor data (spaceborne, airborne, and UAV sensors) in Nambia \cite{Booysen2014}; (b) The trade-off between spectral and spatial resolutions; (c) Elevation information obtained by LiDAR sensors from the University of Houston; (d) Time-series data analysis for assessing the dynamic of changes using RGB and urban images captured from 2001 to 2006 in Dubai.}
\label{fig:FigIntro1}
\end{figure*}

Fortunately, the above increase in the number and heterogeneity of data sources (presenting both challenge and opportunity) has been paralleled by increases in computing power, by efforts to make data more open, available and interoperable, and by advances in methods for data fusion, which are reviewed here \cite{NIELAND201586}. There exist a very wide range of approaches to data fusion (e.g., \cite{GHASSEMIAN201675, LI2017100, LIU2018158}). This paper seeks to review them by class of data modality (e.g., optical, SAR, laser scanning) because methods for these modalities have developed somewhat differently, according to each research community. Given this diversity, it is challenging to synthesize multisource data fusion approaches into a single framework, and that is not the goal here. Nevertheless, a general framework for measurement and sampling processes (i.e., forward processes) is now described briefly to provide greater illumination of the various data fusion approaches (i.e., commonly inverse processes or with elements of inverse processing) that are reviewed in the following sections. Due to the fact that the
topic of multisensor data fusion is extremely broad and that
specific aspects have been reviewed already we have to restrict
what is covered in the manuscript and, therefore, do not
address a few topics such as the fusion of SAR and optical
data.

We start by defining the space and properties of interest. In remote sensing, there have historically been considered to be four dimensions in which information is provided. These are: spatial, temporal, spectral, and radiometric; that is, 2D spatially, 1D temporally, and 1D spectrally with ``radiometric" referring to numerical precision. The electromagnetic spectrum (EMS) exists as a continuum and, thus, lends itself to high-dimensional feature space exploration through definition of multiple wavebands (spectral dimension). LiDAR and TLS, in contrast to most optical and SAR sensors, measure a surface in 3D spatially. Recent developments in photo- and radargrammetry
such as Structure from Motion (SfM) and InSAR,
have increased the availability of 3D data. This expansion of the dimensionality of interest to 3D in space and 1D in time makes image and data fusion additionally challenging \cite{Xia2018LH}. The properties measured in each case vary, with SAR measuring backscatter, optical sensors (including hyperspectral) measuring the visible and infrared parts of the EMS, and laser scanners measuring surface elevation in 3D. Only surface elevation is likely to be a primary interest, whereas reflectance and backscatter are likely to be only indirectly related to the property of interest. 

Secondly, we define measurement processes. A common ``physical model" in remote sensing is one of four component models: scene model, atmosphere model, sensor model, and image model \cite{103293, JOSEPH20009, ZHAO2001202, 5491159, 7855610, 7926323}. The scene model defines the subject of interest (e.g., land cover, topographic surface), while the atmosphere model is a transform of the EMS from surface to sensor, the sensor model represents a measurement process (e.g., involving a signal-to-noise ratio, the point spread function) and the image model is a sampling process (e.g., to create the data as an image of pixels on a regular grid). 

Third, the sampling process implied by the image model above can be expanded and generalized to three key parameters (the sampling extent, the sampling scheme, and the sampling support), each of which has four further parameters (size, geometry, orientation, and position). The support is a key sampling parameter which defines the space on which each observation is made; it is most directly related to the point spread function in remote sensing, and is represented as an image pixel \cite{WANG2017127}. The combination and arrangement of pixels as an image defines the spatial resolution of the image. Fusion approaches are often concerned with the combination of two or more datasets with different spatial resolutions such as to create a unified dataset at the finest resolution \cite{ATKINSON2013106, rs8100797, 8325487}. Fig.~\ref{fig:FigIntro1}(a)  demonstrates schematically the multiscale nature (different spatial resolutions) of diverse datasets captured by spaceborne, airborne, and UAV sensors. In principle, there is a relation between spatial resolution and scene coverage, i.e., data with a coarser spatial resolution (spaceborne data) have a larger scene coverage while data with a finer spatial resolution have a limited coverage (UAV data).

All data fusion methods attempt to overcome the above measurement and sampling processes, which fundamentally limit the amount of information transferring from the scene to any one particular dataset. Indeed, in most cases of data fusion in remote sensing the different datasets to be fused derive in different ways from the same scene model, at least as defined in a specific space-time dimension and with specific measurable properties (e.g., land cover objects, topographic surface). Understanding these measurement and sampling processes is, therefore, key to characterizing methods of data fusion since each operates on different parts of the sequence from scene model to data. For example, it is equally possible to perform the data fusion process in the scene space (e.g., via some data generating model such as a geometric model) as in the data space (the more common approach) \cite{7926323}. 

Finally, we define the ``statistical model" framework as including: (i) measurement to provide data, as described above, (ii) characterization of the data through model fitting, (iii) prediction of unobserved data given (ii), and (iv) forecasting \cite{Mangion2018OnSA}. (i), (ii), and (iii) are defined in space or space-time, while (iv) extends through time beyond the range of the current data. Prediction (iii) can be of the measured property {\bf x} (e.g., reflectance or topographic elevation, through interpolation) or it can be of a property of interest {\bf y} to which the measured {\bf x} data are related (e.g., land cover or vegetation biomass, through classification or regression-type approaches). Similarly, data fusion can be undertaken on {\bf x} or it can be applied to predict {\bf y} from {\bf x}. Generally, therefore, data fusion is applied either between (ii) and (iii) (e.g., fusion of {\bf x} based on the model in (ii)), as part of prediction (e.g., fusion to predict {\bf y}) or after prediction of certain variables (e.g., ensemble unification). In this paper, the focus is on data fusion to predict {\bf x}. 

Data fusion is made possible because each dataset to be fused represents a different view of the same real world defined in space and time (generalized by the scene model), with each view having its own measurable properties, measurement processes, and sampling processes. Therefore, crucially, one should expect some level of coherence between the real world (the source) and the multiple datasets (the observations), as well as between the datasets themselves, and this is the basis of most data fusion methods. This concept of coherence is central to data fusion \cite{5}. 

Attempts to fuse datasets are potentially aided by knowledge of the structure of the real world. The real world is spatially correlated, at least at some scale \cite{10.2307/143141} and this phenomenon has been used in many algorithms (e.g., geostatistical models \cite{5}). Moreover, the real world is often comprised of functional objects (e.g., residential houses, roads) that have expectations around their sizes and shapes, and such expectations can aid in defining objective functions (i.e., in optimization solutions) \cite{ZHANG2018133}. These sources of prior information (on real world structure) constrain the space of possible fusion solutions beyond the data themselves. 

Many key application domains stand to benefit from data fusion processing. For example, there exists a very large number of applications where an increase in spatial resolution would add utility, which is the center of focus in Section II of this paper. These include land cover classification, urban-rural definition, target identification, geological mapping, and so on (e.g., \cite{10.3389/feart.2017.00017}). A large focus of attention currently is on the specific problem that arises from the trade-off in remote sensing between spatial resolution and temporal frequency; in particular the fusion of coarse-spatial-fine-temporal-resolution with fine-spatial-coarse-temporal-resolution space-time datasets such as to provide frequent data with fine spatial resolution \cite{9, 31, 8036397, 33}, which will be detailed in Section II and V of this paper. Land cover classification is one of the most vibrant fields of research in the remote sensing community \cite{Ghamisi-GRSM-2017,8113122}, which attempts to differentiate between several land cover classes available in the scene, can substantially benefit from data fusion. Another example is the trade-off between spatial resolution and spectral resolution (Fig.~\ref{fig:FigIntro1}(b)) to produce fine-spectral-spatial resolution images, which plays an important role for land cover classification and geological mapping. As can be seen in Fig.~\ref{fig:FigIntro1}(b), both fine spectral and spatial resolutions are required to provide detailed spectral information and avoid the ``mixed-pixel" phenomenon at the same time. Further information about this topic can be found in Section II. Elevation information provided by LiDAR and TLS (see Fig.~\ref{fig:FigIntro1}(c)) can be used in addition to optical data to further increase classification and mapping accuracy, in particular for classes of objects, which are made up of the same materials (e.g., grassland, shrubs, and trees). Therefore, Sections III and IV of this paper are dedicated to the topic of elevation data fusion and their integration with passive data. Furthermore, new sources of ancillary data obtained from social media, crowd sourcing, and scraping the internet can be used as additional sources of information together with airborne and spaceborne data for smart city and smart environment applications as well as hazard monitoring and identification. This young, yet active, field of research is the focus of Section VI.

Many applications can benefit from fused fine-resolution, time-series datasets, particularly those that involve seasonal or rapid changes, which will be elaborated in Section V. Fig.~\ref{fig:FigIntro1}(d) shows the dynamic of changes for an area in Dubai from 2001 to 2006 using time-series of RGB and urban images. For example, monitoring of vegetation phenology (the seasonal growing pattern of plants) is crucial to monitoring deforestation \cite{7586093} and crop yield forecasting, which mitigates against food insecurity globally, natural hazards (e.g. earth-
quakes, landslides) or illegal activities such as pollutions (e.g.
oil spills, chemical leakages). However, such information is provided globally only at very coarse resolution, meaning that local smallholder farmers cannot benefit from such knowledge. Data fusion can be used to provide frequent data needed for phenology monitoring, but at a fine spatial resolution that is relevant to local farmers \cite{doi:10.1080/01431161.2017.1395969}. Similar arguments can be applied to deforestation where frequent, fine resolution data may aid in speeding up the timing of government interventions \cite{rs9040381, 7586093}. The case for fused data is arguably even greater for rapid change events; for example, forest fires and floods. In these circumstances, the argument for frequent updates at fine resolution is obvious. While these application domains provide compelling arguments for data fusion, there exist many challenges including: (i) the data volumes produced at coarse resolution via sensors such as MODIS and MERIS are already vast, meaning that fusion of datasets most likely needs to be undertaken on a case-by-case basis as an on-demand service and (ii) rapid change events require ultra-fast processing meaning that speed may outweigh accuracy in such cases \cite{ZHU2017370}. In summary, data fusion approaches in remote sensing vary greatly depending on the many considerations described above, including the sources of the datasets to be fused. In the following sections, we review data fusion approaches in remote sensing according to the data sources to be fused only, but the further considerations introduced above are relevant in each section. 

The remainder of this review is divided into the following sections. First, we review pansharpening and resolution enhancement approaches in Section II. Then, we will move on by discussing point cloud data fusion in Section III. Section IV is devoted to hyperspectral and LiDAR data fusion. Section V presents an overview of multitemporal data fusion. Major recent  advances in big data and social media fusion are presented in Section IV. Finally, Section VII draws conclusions.

\section{Pansharpening and Resolution Enhancement}
\begin{figure}
  \centering
 \centering
\includegraphics[width=0.999\linewidth]{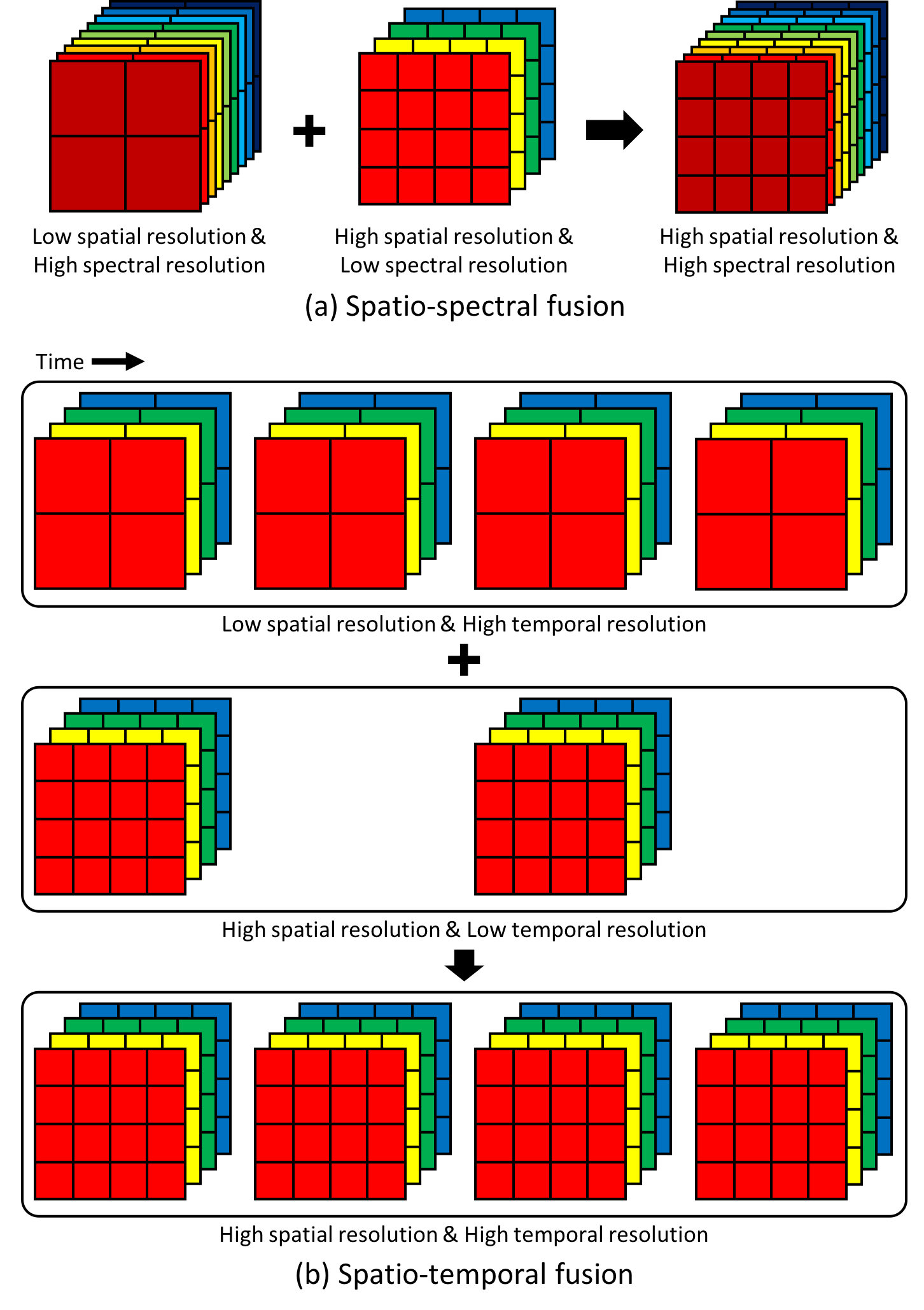}
  \caption{Schematic illustrations of (a) spatio-spectral fusion and (b) spatio-temporal fusion.}
  \label{fig:Sec2}
\end{figure}

Optical Earth observation satellites have trade-offs in spatial, spectral, and temporal resolutions. Enormous efforts have been made to develop data fusion techniques for reconstructing synthetic data that have the advantages of different sensors. Depending on which pair of resolutions has a tradeoff, these technologies can be divided into two categories: (1) spatio-spectral fusion to merge fine-spatial and fine-spectral resolutions [see Fig.~\ref{fig:Sec2}(a)]; (2) spatio-temporal fusion to blend fine-spatial and fine-temporal resolutions [see Fig.~\ref{fig:Sec2}(b)]. This section provides overviews of these technologies with recent advances.

\subsection{Spatio-spectral fusion}

Satellite sensors such as WorldView and Landsat ETM+ can observe the Earth's surface at different spatial resolutions in different wavelengths. For example, the spatial resolution of the eight-band WorldView multispectral image is 2 m, but the single band panchromatic (PAN) image has a spatial resolution of 0.5 m. Spatio-spectral fusion is a technique to fuse the fine spatial resolution images (e.g., 0.5 m WorldView PAN image) with coarse spatial resolution images (e.g., 2 m WorldView multispectral image) to create fine spatial resolution images for all bands. Spatio-spectral fusion is also termed pan-sharpening when the available fine spatial resolution image is a single PAN image. 
When multiple fine spatial resolution bands are available, spatio-spectral fusion is referred to as multiband image fusion, where two optical images with a trade-off between spatial and spectral resolutions are fused to reconstruct fine-spatial and fine-spectral resolution imagery. Multiband image fusion tasks include multiresolution image fusion of single-satellite multispectral data (e.g., MODIS and Sentinel-2) and hyperspectral and multispectral data fusion~\cite{Yokoya17}.

\begin{figure}
  \centering
 \centering
\includegraphics[width=0.999\linewidth]{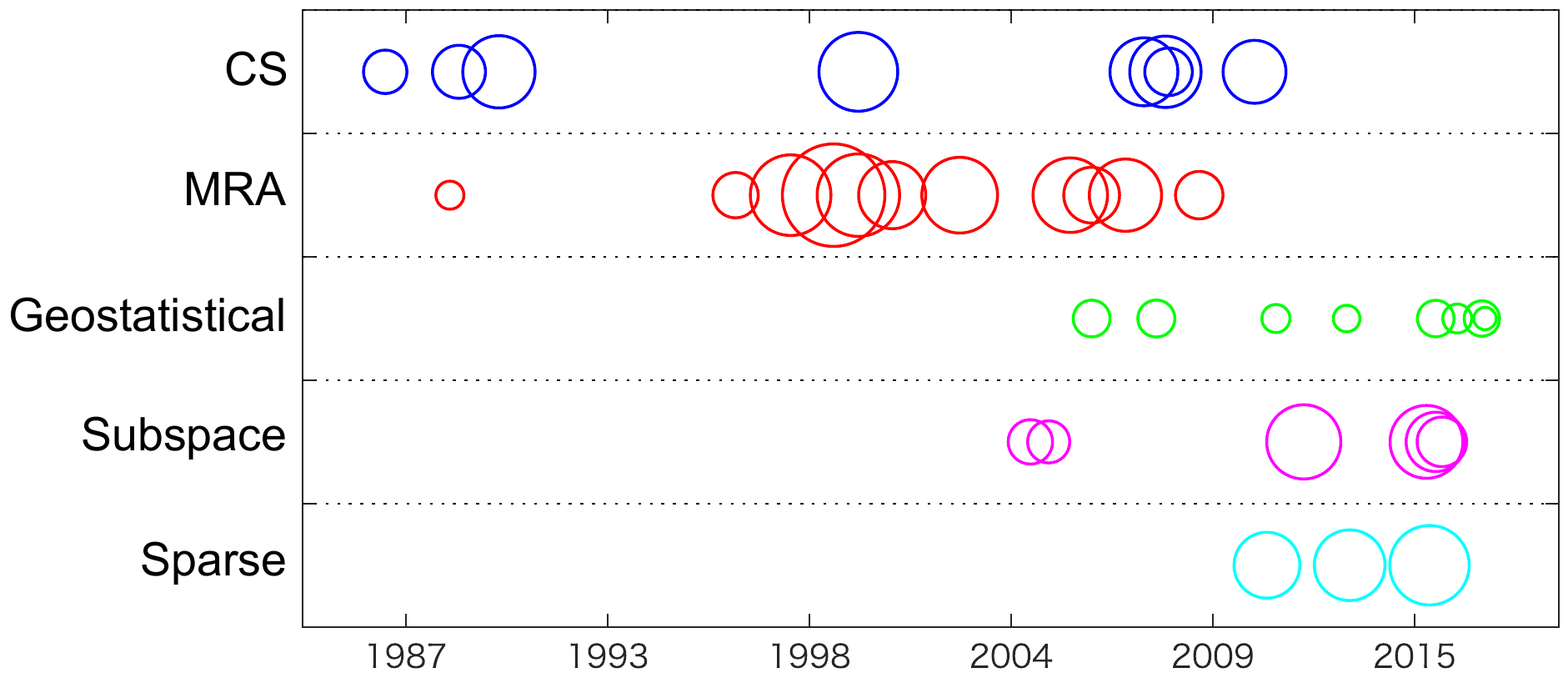}
  \caption{The history of the representative literature of five approaches in spatio-spectral fusion. The size of each circle is proportional to the annual average number of citations. For each category, from left to right, circles correspond to~\cite{welch1987merging,kwarteng1989extracting,Carper90,ranchin2000fusion,laben2000process,aiazzi2002context,Aiazzi07,shah2008efficient,rahmani2010adaptive} for \textit{CS},~\cite{yocky1996multiresolution,zhou1998wavelet,nunez1999multiresolution,Liu00,otazu2005introduction,Aiazzi06,nencini2007remote} for \textit{MRA},~\cite{1,3,2,4},~\cite{5},~\cite{6},~\cite{9},~\cite{8} for \textit{Geostatistical},~\cite{Hardie04,Eismann05,Yokoya12a,Simoes15,Wei15b,Wei15c} for \textit{Subspace}, and~\cite{SLi11,Zhu13,Wei15a} for \textit{Sparse}.}
  \label{fig:DS0}
\end{figure}

Over the past decades, spatio-spectral fusion has motivated considerable research in the remote sensing community. Most spatio-spectral fusion techniques can be categorized into at least one of five approaches: 1) component substitution (CS), 2) multiresolution analysis (MRA), 3) geostatistical analysis, 4) subspace representation, and 5) sparse representation. Fig.~\ref{fig:DS0} shows the history of representative literature with different colors (or rows) representing different categories of techniques. The size of each circle is proportional to the annual average number of citations (obtained by Google Scholar on January 20, 2018), which indicates the impact of each approach in the field. The main concept and characteristics of each category are described below.

\subsubsection{Component Substitution}
CS-based pan-sharpening methods spectrally transform the multispectral data into another feature space to separate spatial and spectral information into different components. Typical transformation techniques include intensity-hue-saturation (IHS)~\cite{Carper90}, principal component analysis (PCA)~\cite{kwarteng1989extracting}, and Gram-Schmidt~\cite{laben2000process} transformations. Next, the component that is supposed to contain the spatial information of the multispectral image is substituted by the PAN image after adjusting the intensity range of the PAN image to that of the component using histogram matching. Finally, the inverse transformation is performed on the modified data to obtain the sharpened image.

Aiazzi et al. (2007) proposed the general CS-based pan-sharpening framework, where various methods based on different transformation techniques can be explained in a unified way~\cite{Aiazzi07}. In this framework, each multispectral band is sharpened by injecting spatial details obtained as the difference between the PAN image and a coarse-spatial-resolution synthetic component multiplied by a band-wise modulation coefficient. By creating the synthetic component based on linear regression between the PAN image and the multispectral image, the performances of traditional CS-based techniques were greatly increased, mitigating spectral distortion.

CS-based fusion techniques have been used widely owing to the following advantages: i) high fidelity of spatial details in the output, ii) low computational complexity, and iii) robustness against misregistration. On the other hand, the CS methods suffer from global spectral distortions when the overlap of spectral response functions (SRFs) between the two sensors is limited.

\subsubsection{Multiresolution Analysis}
As shown in Fig.~\ref{fig:DS0}, great effort has been devoted to the study of MRA-based pan-sharpening algorithms particularly between 2000 and 2010 and they have been used widely as benchmark methods for more than ten years. The main concept of MRA-based pan-sharpening methods is to extract spatial details (or high-frequency components) from the PAN image and inject the details multiplied by gain coefficients into the multispectral data. MRA-based pan-sharpening techniques can be characterized by 1) the algorithm used for obtaining spatial details (e.g., spatial filtering or multiscale transform), and 2) the definition of the gain coefficients. Representative MRA-based fusion techniques are based on box filtering~\cite{Liu00}, Gaussian filtering~\cite{Aiazzi06}, bilateral filtering~\cite{sun2014nearest}, wavelet transform~\cite{nunez1999multiresolution,otazu2005introduction}, and curvelet transform~\cite{nencini2007remote}. The gain coefficients can be computed either locally or globally. 

Selva et al. (2015) proposed a general framework called hypersharpening that extends MRA-based pan-sharpening methods to multiband image fusion by creating a fine spatial resolution synthetic image for each coarse spatial resolution band as a linear combination of fine spatial resolution bands based on linear regression~\cite{Selva15}. 

The main advantage of the MRA-based fusion techniques is its spectral consistency. In other words, if the fused image is degraded in the spatial domain, a degraded image is spectrally consistent with the input coarse-spatial and fine-spectral resolution image. The main shortcoming is that its computational complexity is greater than that of CS-based techniques. 

\subsubsection{Geostatistical Analysis}
Geostatistical solutions provide another family of approaches for spatio-spectral fusion. This type of approach can preserve the spectral properties of the original coarse images. That is, when the downscaled prediction is upscaled to the original coarse spatial resolution, the result is identical to the original one (i.e., perfect coherence). Pardo-Iguzquiza et al. \cite{1} developed a downscaling cokriging (DSCK) method to fuse the Landsat ETM+ multispectral images with the PAN image. DSCK treats each multispectral image as the primary variable and the PAN image as the secondary variable. DSCK was extended with a spatially adaptive filtering scheme \cite{2}, in which the cokriging weights are determined on a pixel basis, rather than being fixed in the original DSCK. Atkinson et al. \cite{3} extended DSCK to downscaled the multispectral bands to a spatial resolution finer than any input images, including the PAN image. DSCK is a one-step method, and it involves auto-semivariogram and cross-semivariogram modeling for each coarse band \cite{4}.

Sales et al. \cite{4} developed a kriging with external drift (KED) method to fuse 250 m Moderate Resolution Imaging Spectroradiometer (MODIS) bands 1-2 with 500 m bands 3-7. KED requires only auto-semivariogram modeling for the observed coarse band and simplifies the semivariogram modeling procedure, which makes it easier to implement than DSCK. As admitted in Sales et al. \cite{4}, however, KED suffers from expensive computational cost, as it computes kriging weights locally for each fine pixel. The computing time increases linearly with the number of fine pixels to be predicted.

Wang et al. \cite{5} proposed an area-to-point regression kriging (ATPRK) method to downscale MODIS images. ATPRK includes two steps: regression-based overall “trend” estimation and area-to-point kriging (ATPK)-based residual downscaling. The first step constructs the relationship between the fine and coarse spatial resolution bands by regression modelling and then the second step downscales the coarse residuals from the regression process with ATPK. The downscaled residuals are finally added back to the regression predictions to produce fused images. ATPRK requires only auto-semivariogram modeling and is much easier to automate and more user-friendly than DSCK. Compared to KED, ATPRK calculates the kriging weights only once and is a much faster method. ATPRK was extended with an adaptive scheme (called AATPRK), which fits a regression model using a local scheme where the regression coefficients change across the image \cite{6}. 
For fast fusion of hyperspectral and multispectral images, ATPRK was extended with an approximate version \cite{8}. The approximate version greatly expedites ATPRK and also has a very similar performance in fusion. ATPRK was also employed for fusion of the Sentinel-2 Multispectral Imager (MSI) images acquired from the recently launched Sentinel-2A satellite. Specifically, the six 20 m bands were downscaled to 10 m spatial resolution by fusing them with the four observed 10 m bands \cite{9}.

\begin{figure*}
  \centering
 \centering
  \includegraphics[width=0.999\linewidth]{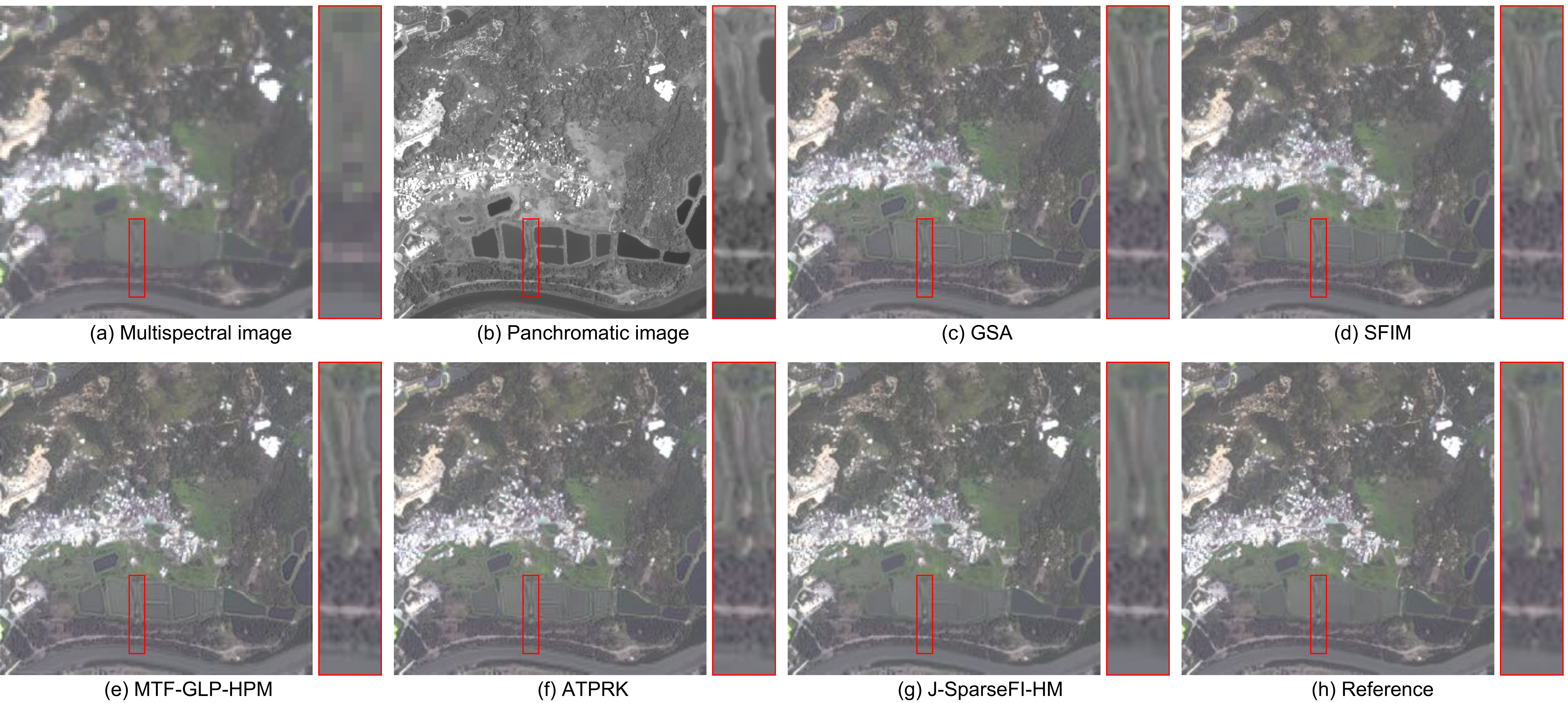}
  \caption{The Hong Kong WorldView-2 dataset (bands 4, 3, and 2 as RGB). (a) 8 m coarse multispectral image, (b) 2 m PAN image (c) GSA, (d) SFIM, (e) MTF-GLP-HPM, (f) ATPRK, (g) J-SparseFI-HM, and (h) 2 m reference image.}
  \label{fig:DS1}
\end{figure*}

\begin{table}[t]
  \centering\addtolength{\tabcolsep}{-2.pt}
  \caption{Quantitative assessment of five representative pan-sharpening methods for the Hong Kong WorldView-2 dataset}
    \begin{tabular}{cccccc}
    \toprule
 Category   & Method	& PSNR & SAM & ERGAS & $Q2^{n}$ \\ \hline
---    &    Ideal	& $\inf$	& 0	& 0	& 1 \\
CS & GSA & 36.9624 & 1.9638 & 1.2816 & 0.86163 \\
MRA &SFIM	& 36.4975 & 1.8866 & 1.2857 & 0.86619 \\
MRA &MTF-GLP-HPM & 36.9298 & 1.8765 & 1.258 & 0.85945 \\
Geostatistical &ATPRK & 37.9239 & 1.7875 & 1.1446 & 0.88082 \\
Sparse &J-SparseFI-HM & 37.6304 & 1.6782 & 1.0806 & 0.88814 \\
\bottomrule
    \end{tabular}%
  \label{tab:DS1}%
\end{table}%

\subsubsection{Subspace Representation}
As indicated in Fig.~\ref{fig:DS0}, research on subspace-based fusion techniques has become very popular recently. Most of these techniques have been developed for multiband image fusion. The subspace-based methods solve the fusion problem via the analysis of the intrinsic spectral characteristics of the observed scene using a subspace spanned by a set of basis vectors (e.g., a principal component basis and spectral signatures of endmembers). The problem is formulated as the estimation of the basis at a fine-spectral resolution and the corresponding subspace coefficients at a fine-spatial resolution. This category of techniques includes various methods based on Bayesian probability~\cite{Wei15b}, matrix factorization~\cite{Yokoya12a}, and spectral unmixing~\cite{Lanaras15}. The interpretation of the fusion process is straightforward in the case of unmixing-based methods: endmembers and their fine-spatial-resolution fractional abundances are estimated from the input images; the output is reconstructed by multiplying the endmember matrix and the abundance matrix.

A recent comparative review on multiband image fusion in~\cite{Yokoya17} demonstrated that unmixing-based methods are capable of achieving accurate reconstruction results even when the SRF overlap between the two sensors is limited. Many subspace-based algorithms are computationally expensive compared to CS- and MRA-based methods due to iterative optimization. Recent efforts for speeding up the fusion procedure~\cite{Wei15c} are key to the applicability of this family of techniques for large-sized images obtained by operational satellites (e.g., Sentinel-2). Another drawback of the subspace-based methods is that they can introduce unnatural artifacts in the spectral domain due to imperfect subspace representations.

\subsubsection{Sparse Representation}
In recent years, spatio-spectral fusion approaches based on patch-wise sparse representation have been developed along with the theoretical development of compressed sensing and sparse signal recovery. Pan-sharpening based on sparse representation can be regarded as a special case of learning-based super-resolution, where correspondence between coarse- and fine-spatial-resolution patches are learned from a database (or a dictionary). Li et al. (2011) proposed the first sparse-representation-based pan-sharpening method that exploits various external fine-spatial-resolution multispectral images as a database~\cite{SLi11}. By considering the PAN image as a source for constructing a dictionary, it is possible to deal with the general problem setting of pan-sharpening, where there is only one pair of PAN and multispectral images is available~\cite{Zhu13}. Sparse representations have been introduced into the subspace-based fusion scheme to regularize fine-spatial-resolution subspace coefficients based on Bayesian probability~\cite{Wei15a}. 

It is noteworthy that sparse-representation-based techniques are capable of sharpening spatial details that are not visible in the fine spatial resolution image at exactly the same location by reconstructing each patch of the output as a linear combination of non-local patches of the fine-spatial-resolution image. The critical drawback is its extremely high computational complexity, sometimes requiring supercomputers to process fusion tasks in an acceptable time. 

We compare five representative pan-sharpening algorithms, namely, GSA~\cite{Aiazzi07}, SFIM~\cite{Liu00}, MTF-GLP-HPM~\cite{vivone2014contrast}, ATPRK~\cite{5}, and J-SparseFI-HM~\cite{Grohnfeldt17} using WorldView-2 data taken over Hong Kong. The original dataset consists of 0.5 m GSD PAN and 2 m GSD 8 multispectral bands. To assess the quality of pan-sharpened images, we adopt Wald's protocol, which degrades the original PAN and multispectral images to 2 m and 8 m GSDs, respectively, with the original multispectral bands being the reference. For quantitative evaluation, we use peak signal-to-noise ratio (PSNR), spectral angle mapper (SAM), {\it erreur relative globale adimensionnelle de synth\`{e}se} (ERGAS)~\cite{Wald00}, and $Q2^n$~\cite{Garzelli09}, which are all well-established quality measures in pan-sharpening. PSNR quantifies the spatial reconstruction
quality of each band, and the SAM index measures the spectral information preservation at each pixel. We use the average PSNR and SAM values. ERGAS and Q2n are global reconstruction indices.

The experimental results are compared both visually and quantitatively in Fig.~\ref{fig:DS1} and Table~\ref{tab:DS1}, respectively. The quality measures in Table~\ref{tab:DS1} are consistent with the literature: GSA, SFIM, and MTF-GLP-HPM provide the competitive baselines, ATPRK clearly outperforms the baselines, and J-SparseFI-HM achieves further increases in accuracy. In Fig.~\ref{fig:DS1}, we can observe different characteristics of the investigated methods. For instance, GSA, SFIM, MTF-GLP-HPM, and ATPRK show sharper edges but also artifacts along object boundaries (e.g., between water and vegetation) where brightness is reversed between the PAN image and each band. J-SparseFI-HM  deals with such situations and produces visually natural results owing to its non-local sparse representation capability.

\subsection{Spatio-temporal fusion}

For remote sensing-based global monitoring, there always exists a trade-off between spatial resolution and temporal revisit frequency (i.e., temporal resolution). For example, the MODIS satellite can provide data on a daily basis, but the spatial resolution (250 m to 1000 m) is often too coarse to provide explicit land cover information, as such information may exist at a finer spatial scale than the sensor resolution. The Landsat sensor can acquire images at a much finer spatial resolution of 30 m, but has a limited revisit capability of 16 days. Fine spatial and temporal resolution data are crucial for timely monitoring of highly dynamic environmental, agricultural or ecological phenomena. The recent development of remotely piloted aircraft systems (RPAS) or drones will
provide a huge amount of multisource data with very high
spatial and temporal resolutions.

Spatio-temporal fusion is a technique to blend fine spatial resolution, but coarse temporal resolution (e.g., Landsat) data and fine temporal resolution, but coarse spatial resolution data to create fine spatio-temporal resolution (e.g., MODIS) data \cite{10,11,12}. Its implementation is performed based on the availability of at least one coarse-fine spatial resolution image pair (e.g., MODIS-Landsat image pair acquired on the same day) or one fine spatial resolution land cover map that is temporally close to the prediction day. Over the past decade, several spatio-temporal fusion methods have been developed and they can generally be categorized into image-pair-based and spatial unmixing-based methods.

The spatial and temporal adaptive reflectance fusion model (STARFM) \cite{13} is one of the earliest and most widely used spatio-temporal fusion methods. It is a typical image-pair-based method. It assumes that the temporal changes of all land cover classes within a coarse pixel are consistent, which is more suitable for homogeneous landscapes dominated by pure coarse pixels. To enhance STARFM for heterogeneous landscapes, an enhanced STARFM (ESTARFM) method was developed \cite{14}. ESTARFM requires two coarse-fine image pairs to estimate the temporal change rate of each class separately and assumes the change rates to be stable during the relevant period \cite{15}. Moreover, some machine learning-based methods were proposed, including sparse representation \cite{16,17}, extreme learning machine \cite{18}, artiﬁcial neural network \cite{19}, and deep learning \cite{20}. These methods learn the relationship between the available coarse-fine image pairs, which is used to guide the prediction of fine images from coarse images on other days.

Spatial unmixing-based methods can be performed using only one fine spatial resolution land cover map. The thematic map can be produced by interpretation of the available fine spatial resolution data \cite{21,22,23} or from other sources such as an aerial image \cite{24} or land-use database \cite{25}. This type of methods is performed based on the strong assumption that there is no land-cover/land-use change during the period of interest. Using a fine spatial resolution land-use database LGN5 \cite{25} or a 30 m thematic map obtained by classification of an available Landsat image \cite{23}, 30 m Landsat-like time-series were produced from 300 m Medium Resolution Imaging Spectrometer (MERIS) time-series to monitor vegetation seasonal dynamics. To maintain the similarity between the predicted endmembers and the pre-defined endmembers extracted from the coarse data, Amorós-L\'{o}pez et al. \cite{21, 22} proposed to include a new regularization term to the cost function of the spatial unmixing. Wu et al. \cite{26} and Gevaert et al. \cite{27} extended spatial unmixing to cases with one coarse-fine image pair available. The method estimates changes in class endmember spectra from the time of the image pair to prediction before adding  them to the known fine spatial resolution image. Furthermore, Huang and Zhang \cite{28} developed an unmixing-based spatio-temporal reflectance fusion model (U-STFM) using two coarse-fine image pairs. In addition, the image-pair-based and spatial unmixing-based methods can also be combined \cite{29,30,31}.

Spatio-temporal fusion is essentially an ill-posed problem involving inevitable uncertainty, especially for predicting abrupt changes and heterogeneous landscapes. To this end, Wang et al. \cite{32} proposed to incorporate the freely available 250 m MODIS images into spatio-temporal fusion. Compared to the original 500 m MODIS data, the 250 m data can provide more information for the abrupt changes and heterogeneous landscapes than, and thus, can increase the accuracy of spatio-temporal fusion predictions.

Blending MODIS and Landsat has been the most common spatio-temporal fusion problem over the past decade. Recently, Sentinel-2 and Sentinel-3 are two newly launched satellites for global monitoring. The Sentinel-2 MSI and Sentinel-3 Ocean and Land Colour Instrument (OLCI) sensors have very different spatial and temporal resolutions (Sentinel-2 MSI sensor 10 m, 20 m and 60 m, 10 days, albeit 5 days with 2 sensors, conditional upon clear skies; Sentinel-3 OLCI sensor 300 m, $<$1.4 days with 2 sensors). Wang et al. \cite{33} proposed a new method, called Fit-FC, for spatio-temporal fusion of Sentinel-2 and Sentinel-3 images to create nearly daily Sentinel-2 images. Fit-FC is a three-step method consisting of regression model fitting (RM fitting), spatial filtering (SF) and residual compensation (RC). The Fit-FC method can be implemented using only one image pair and is particularly relevant for cases involving strong temporal changes.

\subsection{Challenges and trends of downscaling}
The major remaining issue in the field of spatio-spectral fusion is how to conduct fair comparisons. Many researchers use their own simulated datasets, and the source code is rarely released. To fairly evaluate the performance of each algorithm, it is necessary to develop benchmark datasets that can be accessible for everyone and include various scenes. Also, it is always desirable to release the source code of each method for enabling reproducible research. In several review papers, researchers have attempted to evaluate many methods with common datasets and to disclose their source code, which is an excellent contribution to the community. However, the diversity of the studied scenes may not be enough to evaluate generalization ability, and also those datasets are not freely available due to a restricted data policy of the original sources. Regarding the source code, there are still many research groups who never release their source code, while always outperforming state-of-the-art algorithms in their papers. It is an urgent issue of the community to arrange benchmark datasets on a platform like the GRSS Data and Algorithm Standard Evaluation (DASE) website~\cite{dase} so that everyone can fairly compete for the performance of the algorithm.

With respect to spatio-temporal fusion, the main challenges lie in the reconstruction of land cover changes and eliminating the differences between coarse and fine spatial resolution time-series. Due to the large difference in the spatial resolution between coarse and fine spatial resolution time-series (e.g., a ratio of 16 for MODIS-Landsat), the prediction of land cover changes (especially for abrupt changes) from coarse images always involves great uncertainty. Most of the existing methods are performed based on the strong assumption of no land cover change, such as the classical STARFM, ESTARFM, and the spatial unmixing-based method. Furthermore, due to the differences in characteristics of sensors, atmospheric condition, and acquisition geometry, the available coarse and fine spatial resolution data (e.g., MODIS and Landsat data) are always not perfectly consistent. The uncertainty is directly propagated to the spatio-temporal fusion process. In future research, it will be of great interest to develop more accurate methods to account for the land cover changes and inconsistency between coarse and fine spatial resolution time-series.

\begin{figure}
  \centering
 \centering
\includegraphics[width=0.999\linewidth]{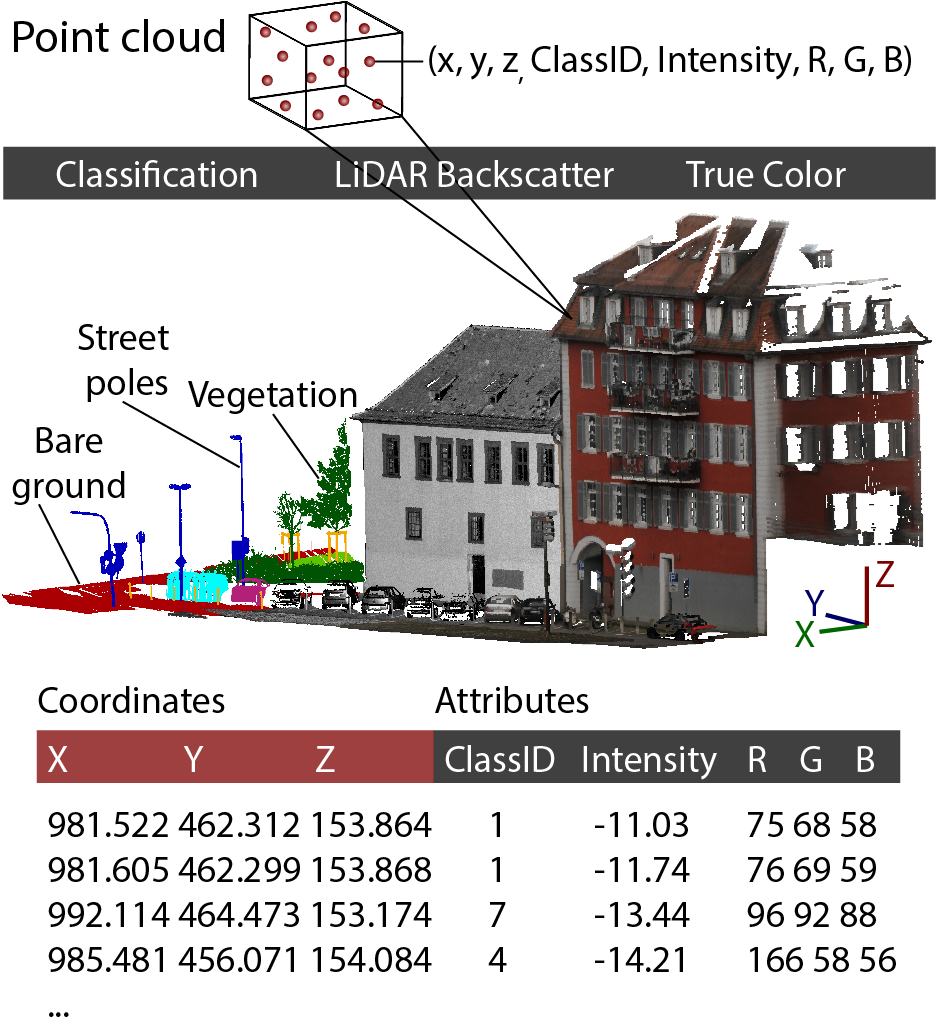}
  \caption{Point cloud data model with the additional point features classification (ID per object class), intensity (LiDAR backscatter information), and true color (RGB values). Each point vector of the point cloud is stored in a table with its 3D coordinate and additional columns per attribute contained in the point cloud.}
  \label{fig:LiDAR1}
\end{figure}
\section{Point cloud data fusion}
Georeferenced point clouds have gained importance in recent years due to a multitude of developments in technology and research that increased their availability (e.g., hardware to capture 3D point clouds) and usability in applications (e.g,. algorithms and methods to generate point clouds and analyze them) \cite{BH7542155}. Research and development with point cloud data is driven from several disciplines (e.g., photogrammetry, computer science, geodesy, geoinformatics, and geography), scientific communities (e.g., LiDAR, computer vision, and robotics) and industry \cite{BHTF}. Spatial and temporal scales to utilize point clouds range from episodic country-wide, large-scale topographic mapping to near real-time usage in autonomous driving applications. Sensors and methods, respectively, to derive point clouds include predominantly LiDAR and photogrammetry \cite{PHOR:PHOR12063}. A further very recent data source of point clouds in research is tomographic SAR \cite{BH6573417}. Also low-cost depth cameras are used increasingly \cite{Hammerle2017}. LiDAR, also referred to as laser scanning, is the only widely used method that records 3D points directly as an active remote, and also close-range, sensing technique \cite{Shan2009}.

\subsection{Point cloud data model}
Although the above-mentioned aspects draw a very broad picture, the common denominator is the point cloud data model, which is the initial data model shared by all multi-source fusion methods that include point clouds. Otepka et al. \cite{ijgi2041038} defined  the georeferenced point cloud data model as a set of points, $\textbf{P}_i$, $i = 1,...,n$, in three-dimensional Cartesian space that is related to a geospatial reference system (e.g., UTM). $\textbf{P}_i$ has at least three coordinates \((x_i,y_i,z_i)^T \in  \rm I\!R^3\) for its position and it can have additional point features, also referred to as attributes \(a_{j,i}\), with \(j = 1,...,m_i\) as the number of point features of point $i$. A point feature, $a_j$, could be the color of a spectral band, LiDAR, or SAR backscatter value, ID of classification or segmentation, local surface normal vector component (e.g., $n_x$, $n_y$, $n_z$), and so forth. Fig.~\ref{fig:LiDAR1} visualizes a point cloud with further point features stored in additional columns of a table with the 3D coordinates. Such point features can originate from the measurement process (e.g., LiDAR intensity \cite{EITEL2016372}), or they can be derived by data post-processing (e.g., segmentation) and fusion with other data sources. Please refer to \cite{ijgi2041038} and \cite{f7090198} for a more detailed description of LiDAR point cloud features. A point in a point cloud, $\textbf{P}_i$, is a vector, \((x_i,y_i,z_i,a_{1,i},....,a_{m_{i},i})^T\), of dimension 3 + $m_i$ with the 3D coordinates as the first three dimensions (see Fig. \ref{fig:LiDAR1}). Generally, the point cloud model supports a variable number of point features $m_i$ and leaves the 3D spatial distribution of \((x_i,y_i,z_i)^T\) up to the point cloud generation process. The main challenges of the point cloud model for fusion with other data sources is the unstructured three-dimensional spatial nature of \textbf{P} and that often no fixed spatial scale and accuracy exist across the dataset. Local neighborhood information must be derived explicitly, which is computationally intensive, and the definition of neighborhood depends on the application and respective processing task \cite{FILIN200671,ijgi2041038}.

\subsection{Concepts of point cloud fusion}
\begin{figure*}
  \centering
 \centering
\includegraphics[width=0.999\linewidth]{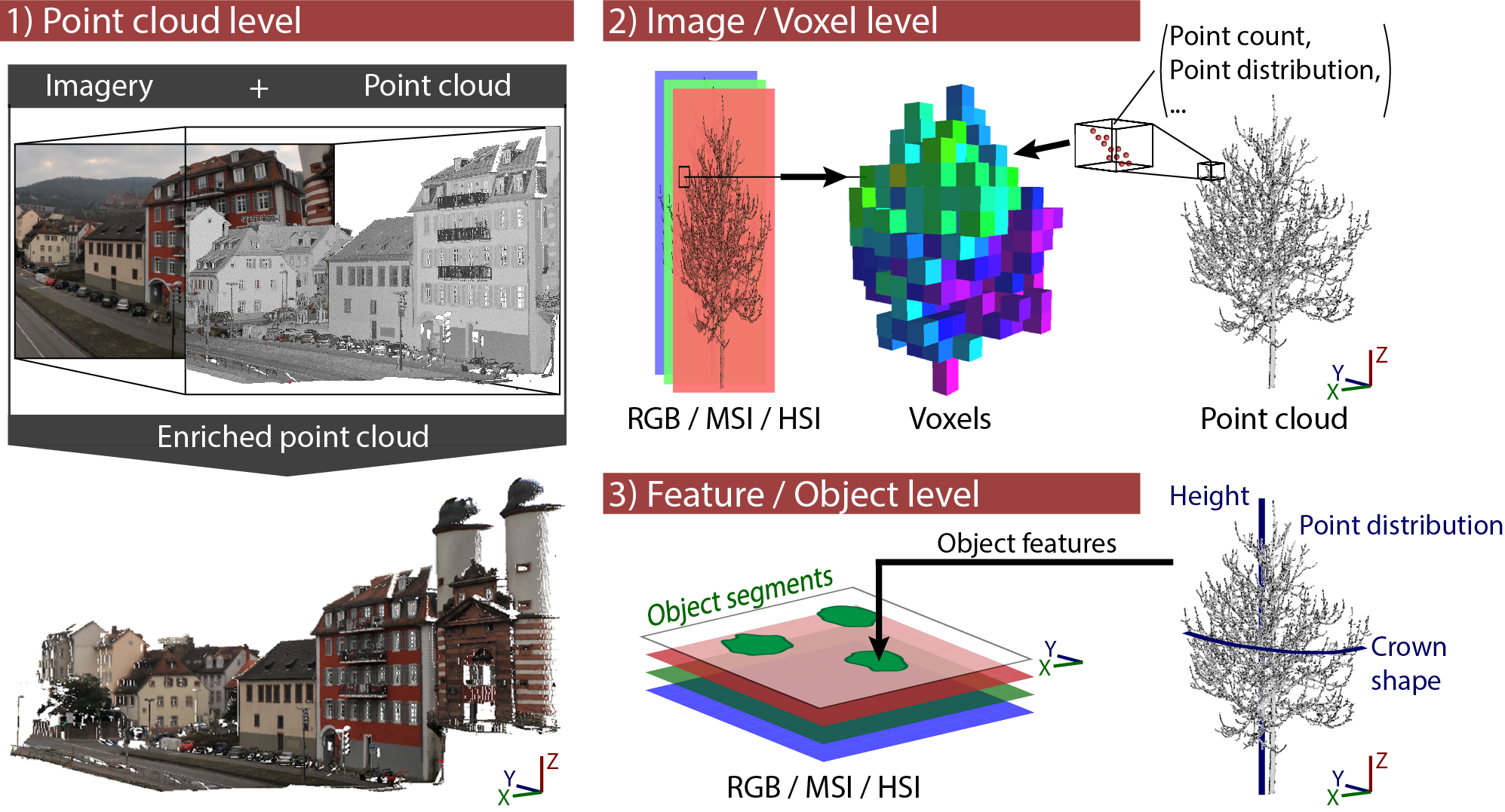}
  \caption{Strategies of point cloud data fusion on (1) point cloud level, (2) image/voxel level, and (3) feature/object level. 1) Visualizes the enrichment of the initial point cloud colored by LiDAR intensity with RGB information from imagery with the RGB-colored point cloud as product. 2) Depicts a voxel model where each voxel contains information from a set of RGB and hyperspectral image layers as well as 3D point cloud features within each voxel. 3) Shows the assignment of features derived from the 3D point cloud to object segments created from raster image data.}
  \label{fig:LiDAR2}
\end{figure*}

The main objectives of point cloud data fusion are to make use of the three-dimensional geometric, spatial-structural and LiDAR backscatter information inherent in point clouds and combine it with spectral data sources or other geoinformation layers, such as GIS data. Zhang and Lin \cite{BHTF} gave a broad overview of applications involving the fusion of optical imagery and LiDAR point clouds. Looking more specifically at the methodology of fusion, three main methodological concepts can be distinguished in the literature with respect to the target model of multi-source point cloud fusion. The target data model of data fusion also determines which methods and software (e.g., image or point cloud processing) are primarily applied to classify the datasets. Based on the target data model (``product") we separate the following strategies (see Fig. \ref{fig:LiDAR2}):
\begin{enumerate}
\item Point cloud level: Enrich the initial point cloud $\textbf{P}$ with new point features.
\item Image/Voxel level: Derive new image layers representing 3D point cloud information.
\item Feature level: Fusion of point cloud information on the segment/object level.
\end{enumerate}

\subsubsection{Point cloud level - Pixel to point and point to point}
Texturing point clouds with image data is a standard procedure for calibrated multi-sensor LiDAR systems for which the transformation from image to point cloud is well-known from lab calibration, such as LiDAR systems with integrated multispectral cameras. For point clouds from photogrammetry - structure-from-motion and dense image matching - the spectral information is already given for each 3D point reconstructed from multiple 2D images \cite{PHOR:PHOR12063}. Thus, the resulting point cloud $\textbf{P}_i$ contains the respective pixel values from the images (e.g., R, G, B) as point features and can be used for classification and object detection. 

The labels of classified hyperspectral data can be transfered to the corresponding 3D points from LiDAR using precise co-registration. With this approach, Buckley et al. \cite{BUCKLEY2013249} related the spectra from close-range hyperspectral imaging pixels to terrestrial LiDAR point clouds to classify inaccessible geological outcrop surfaces. This enables improved visual inspection, but no joint 3D geometric and hyperspectral classification is conducted. A joint classification is presented by Vo et al. \cite{7326746}, in a paper of the 3D-competition of the 2015 IEEE GRSS Data Fusion Contest \cite{BH7542155}. They focused on LiDAR point clouds and RGB images and developed an end-to-end point cloud processing workflow. The authors made use of the colored point cloud and applied a supervised single-point classification (decision tree) to derive the target classes ground, building, and unassigned. This step was followed by the region growing segmentation of the classified ground points to delineate roads. The point features of $\textbf{P}$ were height, image intensity (RGB and HSV), laser intensity, height variation, surface roughness, and normal vector. RGB and laser intensity data particularly supported the exclusion of grass areas and joint classification increased the accuracy of a LiDAR-only solution by 2.3\%. Generally, the majority of published approaches of multi-source point cloud classification, which resulted in a classified point cloud, worked in the image domain and then transfered back the classification results to the point cloud \cite{3dor.20171047}. This allows the use of fast and established image processing, but limits the methods to single point classification because the 3D point neighborhood information is not available in the classification procedure in the image domain, such as it is, for example, in point cloud segmentation.

Point cloud-to-point cloud data fusion is known as point cloud (co-)registration or alignment. Co-registration of point clouds from the same sensor (e.g., within one LiDAR scanning campaign) is a standard pre-processing step in surveying with LiDAR from ground-based and airborne platforms \cite{Shan2009}. Data fusion can be performed by different algorithms, such as point-based [e.g., Iterative Closest Point (ICP)], keypoint-based [e.g., Scale Invariant Feature Transform (SIFT)] or surface-based (e.g., local planes) or any combination \cite{isprs-annals-II-3-57-2014}. This fusion principle is generally valuable if point clouds from different sensor types are merged, which have different accuracies, spatial coverages, and spatial scales as well as being captured at different timestamps.

An image-based 2D registration for merging airborne and multiple terrestrial LiDAR point clouds was used by Paris et al. \cite{7907286} to assess tree crown structures. They used the respective canopy height models for the registration, which was finally applied to the point cloud datasets to derive a fused point cloud. 

A combination of datasets from different methods (e.g., LiDAR and photogrammetry) and platforms can lead to more accurate results compared to the individual use of a source. This was concluded in \cite{10.3389/fpls.2017.02144}, where datasets from two different methods (LiDAR and photogrammetry) and three different platforms [a ground-based platform, a small unmanned aerial systems (UAS)-based platform, and a manned aircraft-based platform] were explored. They merged point clouds from UAS LiDAR, airborne manned LiDAR, and UAS photogrammetry spatially to a single point cloud to estimate the accuracy of bare earth elevation, heights of grasses, and shrubs.

\subsubsection{Image/Voxel level - Point-to-pixel/voxel}
This concept transforms point cloud information into 2D images or voxels that can be analyzed by image processing approaches. In general, a multitude of images can be derived from rich point clouds that derive from point cloud geometry, (LiDAR) backscatter, and also full-waveform LiDAR data directly. Those image bands usually represent elevation, geometric features (e.g., vertical distribution of points within a pixel), and LiDAR intensity-derived features. Ghamisi and H\"ofle \cite{Ghamisi-LiDAR-GRSL-2017} outlined several features that can be derived from LiDAR point clouds to encapsulate the 3D information into image bands for image classification, such as laser echo ratio, variance of point elevation, plane fitting residuals, and echo intensity. The fusion approach of LiDAR and HSI and classification of an urban scene is presented in Section IV. The experiment compares classification results to accuracies of the individual use of HSI.

 A pixel-based convolutional neural network (CNN) was used to perform semantic labeling of point clouds by Boulch et al. \cite{3dor.20171047} based on RGB and geometric information (e.g., depth composite image). Every 3D point is labeled by assigning the derived pixel-wise label predictions to the single 3D points via back projection. The study could apply it to both terrestrial LiDAR and photogrammetric point clouds. 

A fusion of UAV-borne LiDAR, multispectral, and hyperspectral data was presented by Sankey et al. \cite{SANKEY201730} for forest vegetation classification. Furthermore, they used terrestrial LiDAR as reference dataset. The HSI was pre-classified with the mixture-tuned matched filtering subpixel classification technique. The multi-source fusion of UAV LiDAR and hyperspectral data (12 cm GSD) was performed via a decision tree classification approach. The fusion-based result achieved higher user's accuracy for most target classes and also overall accuracy with an increase from 76\% with only HSI to 88\% for HSI and LiDAR data inputs. The largest increase by adding LiDAR was given for vegetation classes that separate well in height.

The combination of the geometric quality of LiDAR and spectral information was used by Gerke and Xiao \cite{GERKE201478} to detect buildings, trees, vegetated ground, and sealed ground. They developed a method to fuse airborne LiDAR and multispectral imagery with two main consecutive steps: 1) Point cloud segmentation (region growing) and classification (mean shift) using 3D LiDAR and spectral information (NDVI/Saturation), 2) supervised (Random Trees) or unsupervised classification - by a Markov random field framework using graph-cuts for energy optimization - of voxels. The voxels contain features derived from 3D geometry and from the spectral image, as well as the results from the initial segmentation step. The results showed that spectral information supported the separation of vegetation from non-vegetation, but shadow areas still caused problems. Point cloud segmentation is sensitive to the color information that was also used in this process, which sometimes led to planes being missed out.

Airborne hyperspectral imagery was combined with full-waveform LiDAR data by Wang and Glennie \cite{WANG20151} to classify nine target land-cover classes (e.g., trees, bare ground, water, asphalt road, etc.). The main goal was to generate synthetic vertical LiDAR waveforms by converting the raw LiDAR waveforms into a voxel model (size of $1.2~\textnormal{m}\times1.2~\textnormal{m}\times0.15~\textnormal{m}$). The voxels were then used to derive several raster features from the vertical distribution of backscatter intensity along the vertical voxels corresponding to one image pixel, and also metrics such as the height of the last return, penetration depth, and maximum LiDAR amplitude. In addition to these raster features, they derived principal components from the original 72 HSI bands and stacked them with the LiDAR features for classification. The fusion of LiDAR waveform data and HSI could increase the overall accuracy using a support vector machine (SVM) classification to 92.61\% compared to 71.30\% using only LiDAR and 85.82\% using only HSI data.

\subsubsection{Feature/Object level}
This concept is based on the previous concepts in terms of data model, which is used to derive objects followed by a classification step. Image or point cloud segmentation, and combined pixel- and object-based approaches can be applied \cite{doi:10.1080/01431161.2015.1015657} to derive the entities for classification. 

With airborne LiDAR images and full-waveform point cloud data, only one data source but two different data models for object-based urban tree classification were used by H\"ofle et al. \cite{Hofle_ISPRS_veg}. They introduced a method to produce segments based on LiDAR point cloud-derived images [e.g., normalized DSM (nDSM) and echo ratio images]. The output segments were enriched by a multitude of geometric and full-waveform features that were computed directly in the 3D point clouds of each segment (e.g., mean echo width). In particular, the geometric 3D point cloud features (e.g., echo ratio) played an important role for vegetation classification because they encapsulated the 3D structure of vegetation. Alonzo et al. \cite{ALONZO201470} also worked at the single tree/crown object level and added HSI to the airborne LiDAR dataset to map urban tree species. They applied canonical variates in a linear discriminant analysis classifier to assign tree species labels to the segments, which were derived from the LiDAR canopy height model. Their LiDAR point cloud-derived structural variables included, for example, median height of returns in crown, average intensity below median height, and so forth. Saarinen et al. \cite{isprs-archives-XLII-3-W3-171-2017} went one step further and fused UAV-borne LiDAR and HSI for mapping biodiversity indicators in boreal forests. After tree crown delineation by watershed segmentation, they derived point cloud-based segment features (e.g., height percentiles and average height) and also spectral segment features (e.g., mean and median spectra). By using nearest-neighbor estimation, the variables of diameter at breast height, tree height, health status, and tree species were determined for each crown segment. In the second step, the biodiversity indicators - structural complexity, amount of deciduous, and dead trees - were derived using single tree variables as input.

Considering multiple sensors, hyperspectral and LiDAR data were fused in an approach proposed by Man et al. \cite{doi:10.1080/01431161.2015.1015657} for urban land-use classification (15 classes) with a combined pixel and feature-level method. LiDAR point cloud information was encapsulated in image layers. Furthermore, they aimed at assessing the contribution of LiDAR intensity and height information, particularly for the classification of shadow areas. Their methodology included pixel-based features such as the nDSM and intensity image from LiDAR, and the inverse minimum noise fraction rotation (MNF) bands, NDVI, and texture features (GLCM) of HSI data. The derived features were input to a supervised pixel-based classification (SVM and maximum likelihood classifiers). Additionally, an edge-based segmentation algorithm was used to derive segments based on LiDAR nDSM, intensity and NDVI images, which was followed by a rule-based classification of the derived objects. The classification outputs of the pixel- and object-based methods were merged by GIS raster calculation. The combination of HSI and LiDAR increased overall accuracy by 6.8\% (to 88.5\%) compared to HSI classification alone. The joint pixel and object-based method increased the overall accuracy by 7.1\% to 94.7\%.

HSI and airborne LiDAR data were used as complementary data sources for crown structure and physiological tree information by Liu et al. \cite{LIU2017170} to map 15 different urban tree species. First, crowns were segmented by watershed segmentation of the canopy height model. Second, LiDAR and hyperspectral features were extracted for the crown segments for the subsequent segment-based random forest classification. The 22 LiDAR-derived crown structural features per segment included, for example, crown shape, laser return intensity, laser point distribution, etc.. They concluded that the combination of LiDAR and HSI increased the single-source classification up to 8.9\% in terms of overall accuracy.

A complex fusion strategy for LiDAR point cloud and HSI image data in a two-stage neural network classification was developed by Rand et al. \cite{doi:10.1117/1.OE.55.7.073101}. First, spectral segmentation of the HSI data was performed by a stochastic expectation-maximization algorithm and spatial segmentation of the LiDAR point cloud with a combined mean-shift and dispersion-based approach. Second, the resulting segments from LiDAR and HSI data were input to a supervised cascaded neural network to derive the final object class labels. The final fusion classification map was produced in 3D by using the elevation values from the LiDAR point cloud. Their approach resulted in a large increase in overall classification accuracy by multi-source fusion (HSI and LiDAR) to 98.5\%, compared to 74.5\% overall accuracy with HSI input only.

\subsection{Challenges and trends of point cloud fusion}
Generally, we can see a large gain in the importance of point clouds. Multi-source fusion including point clouds is already used in a huge variety of fields of applications (see \cite{BHTF}) and reveals several trends:
\begin{itemize}
\item  The increasing use of machine learning methods including point clouds or point cloud derivatives.
\item  The majority of current approaches transform and encapsulate 3D point cloud information into 2D images or voxels and perform fusion and analysis on images or objects. Derived classification labels are transfered back to points afterwards.
\item  The fusion (or joint use) of spectral and 3D point cloud information from single-source photogrammetry (structure-from-motion and dense image matching). The link between point clouds and images is already given via several methodologies.
\item  The fusion of geometric and backscatter point cloud information from LiDAR exhibits increases in terms of classification accuracy.
\end{itemize}

Future research on multi-source fusion with point clouds will need to address the combination of point clouds from different sources and with strongly heterogeneous characteristics (e.g., point density and 3D accuracy). So far, mainly one source of point clouds is used in the fusion process, e.g., the joint use of HSI and LiDAR point clouds. Multispectral \cite{7829363} and even hyperspectral LiDAR data \cite{6419760} offer new possibilities for the fusion of point clouds, as well as of point clouds with MSI/HSI data. The availability of 3D point cloud time-series \cite{EITEL2016372} will also enable investigation of how temporal aspects need to be addressed in fusion and classification approaches.

The number of contributions on HSI and LiDAR rasterized data fusion in the remote sensing community is fast-growing due to the complementary nature of such multi-sensor data. Therefore, Section~\ref{HSILiDAR} is specifically dedicated to the fusion of HSI and LiDAR-derived features to provide readers with an effective review of such fusion schemes.

\section{Hyperspectral and LiDAR}
\label{HSILiDAR}

The efficacy of LiDAR, which is characterized as an active remote sensing technique, for the classification of complex areas (e.g., where many classes are located close to each other) is limited by the lack of spectral information. On the other hand, hyperspectral sensors, which are characterized as passive remote sensing techniques, provide rich and continuous spectral information by sampling the reflective portion of the electromagnetic spectrum, ranging from the visible region (0.4-0.7$\,\mu m$) to the short-wave infrared region (almost 2.4$\,\mu m$) in hundreds of narrow contiguous spectral channels (often 10 nm wide). Such detailed spectral information has made HSIs a valuable source of data for complex scene classification. Detailed and systematic reviews on hyperspectral data classification for characterizing complex scenes have been published in \cite{BenediktssonGhamisiBOOK,Ghamisi-GRSM-2017}. However, HSIs do not contain any information about the elevation and size of different materials, which imposes an inevitable constraint to classify objects that are made up of similar materials (e.g.,  grassland, shrubs, and trees). The aforementioned limitations and capabilities of each sensor, as discussed earlier in the introduction part, have provided the main motivation for fusing HSI and LiDAR.  

The joint use of LiDAR and HSI has already been investigated for diverse applications such as rigorous illumination correction \cite{LH-illuminationcorrection-2017} and quantifying riparian habitat structure \cite{Hall}. However, the main application of this multi-sensor fusion technique is dedicated to scene classification, which is also the pre-eminent focus of this section.

\begin{figure}
  \centering
 \centering
\includegraphics[width=0.95\linewidth]{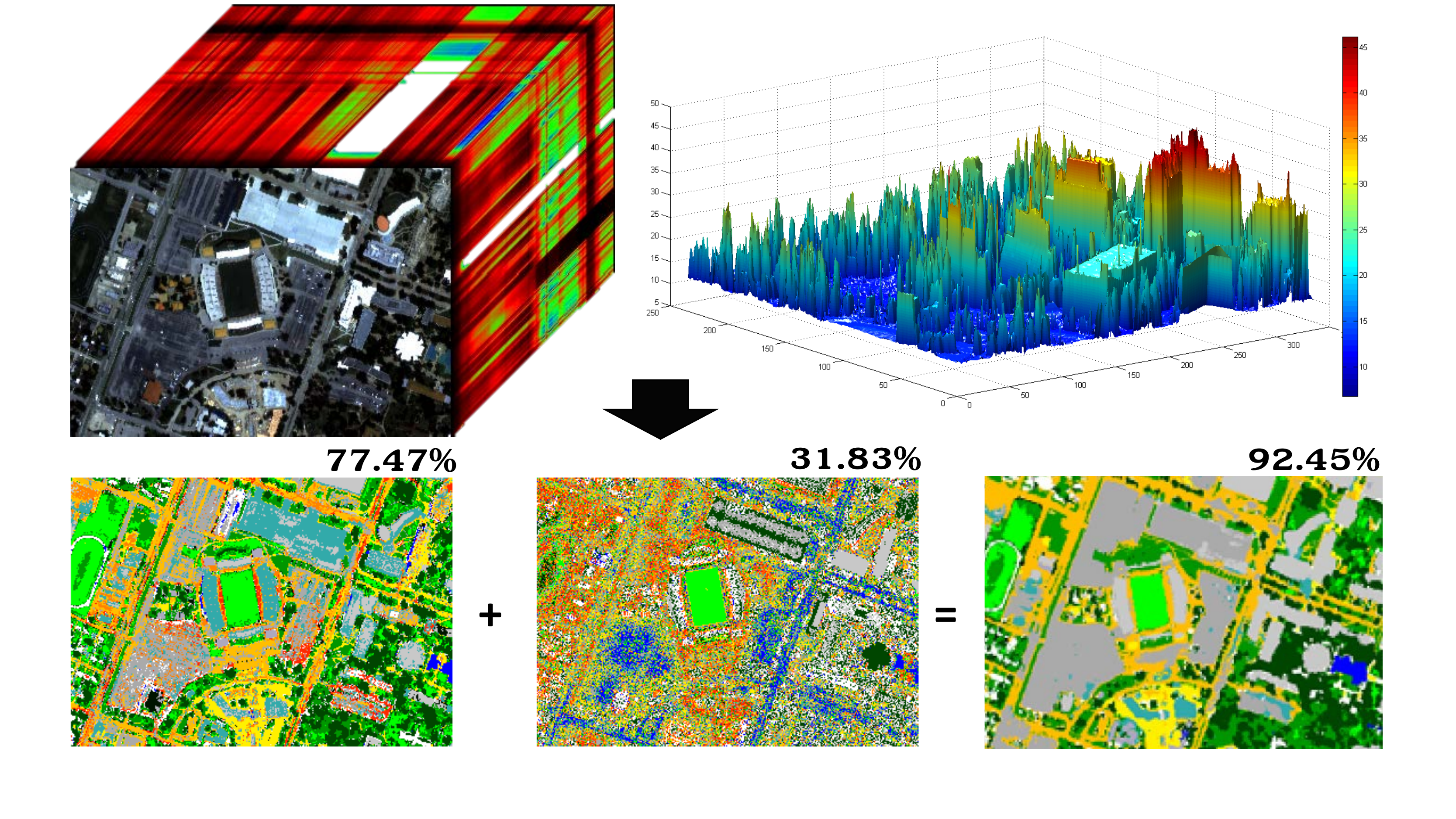}
  \caption{HSI and LiDAR fusion. This figure was generated based on some studies in \cite{OTVCAfusion} where the overall classification accuracy of HSI (77.47\%) and LiDAR (31.83\%) is significantly increased to 92.45\% using a feature fusion approach.}
  \label{MultiModal}
\end{figure}

Several studies such as \cite{YokoyaLH2015, Dalponte-HL-2008} investigated the differentiation of diverse species of trees in complex forested areas, while several other approaches dealt with complex urban area classification (e.g., \cite{HLcomp2013}). Co-registered LiDAR and HSI data were introduced in \cite{2016SPIELiDARHSI}. Fig.~\ref{MultiModal} demonstrates schematically that the fusion of HSI and LiDAR can increase the classification accuracy above that of each individual source considerably (i.e., this figure was generated based on some studies in \cite{OTVCAfusion}).

Below, we discuss briefly a few key approaches for the fusion of LiDAR and HSI, which are categorized in four subsections: Filtering approaches, low-rank models, composite kernels, and deep learning-based fusion approaches. Corresponding to each section, some numerical classification results obtained from the CASI Houston University data (details below) are reported in Table~\ref{tab:HSILiDAR}. To obtain a better numerical evaluation, the classification accuracies of the individual use of HSI obtained by random forest ($\textbf{RF}_{\textnormal{HSI}}$), support vector machine ($\textbf{SVM}_{\textnormal{HSI}}$), and convolutional neural network ($\textbf{CNN}_{\textnormal{HSI}}$) are also listed in Table~\ref{tab:HSILiDAR}.
\begin{table*}[htbp]\scriptsize
  \centering\addtolength{\tabcolsep}{-4.6pt}
  \caption{Houston - The classification accuracy values achieved by different state-of-the-art approaches. The indexes, average accuracy (AA) and overall accuracy (OA), are reported in percentages while the kappa coefficient (\textit{K}) is of no unit.}
    \begin{tabular}{ccccc|cccccccc}
    \toprule
          &       & \multicolumn{3}{c|}{\textbf{Spectral}} & \multicolumn{8}{c}{\textbf{Multisensor fusion}} \\
    \midrule
    \multicolumn{1}{c|}{\textbf{Class name}} & \multicolumn{1}{c|}{\textbf{Train./Test}} & $\textbf{RF}_{\textnormal{HSI}}$ & $\textbf{SVM}_{\textnormal{HSI}}$ & $\textbf{CNN}_{\textnormal{HSI}}$ & $\textbf{EP}_{\textnormal{HSI+LiDAR}}$ & \textbf{GBFF}\cite{HLcomp2013} & \textbf{FFCK}\cite{Zhang-RSL-2017} & \textbf{MLR\textit{sub}}\cite{MkhodadadzadehLH} & \textbf{ALWMJ-KSRC}\cite{JSRC_2016} & \textbf{CNNGBFF}\cite{HL-CNN-JSTARS-2017} & \textbf{SLRCA}\cite{SLRCAfusion} & \textbf{OTVCA}\cite{OTVCAfusion} \\
    \multicolumn{1}{c|}{\textbf{Grass Healthy}} & \multicolumn{1}{c|}{\textbf{198/1053}} & 83.38 & 83.48 & 82.24 & 78.06 & 82.53 & 81.39 & 82.91 & 98.36 & 78.73 & 81.58 & 80.63 \\
    \multicolumn{1}{c|}{\textbf{Grass Stressed}} & \multicolumn{1}{c|}{\textbf{190/1064}} & 98.40 & 96.43 & 98.31 & 84.96 & 98.68 & 99.91 & 81.48 & 98.59 & 94.92 & 99.44 & 99.62 \\
    \multicolumn{1}{c|}{\textbf{Grass Synthetis}} & \multicolumn{1}{c|}{\textbf{192/505}} & 98.02 & 99.80 & 70.69 & 100.00 & 100   & 100   & 100   & 100   & 100   & 98.61 & 100.00 \\
    \multicolumn{1}{c|}{\textbf{Tree}} & \multicolumn{1}{c|}{\textbf{188/1056}} & 97.54 & 98.77 & 94.98 & 95.45 & 98.96 & 97.92 & 95.83 & 98.04 & 99.34 & 96.12 & 96.02 \\
    \multicolumn{1}{c|}{\textbf{Soil}} & \multicolumn{1}{c|}{\textbf{186/1056}} & 96.40 & 98.11 & 97.25 & 98.76 & 100   & 100   & 99.05 & 93.15 & 99.62 & 99.72 & 99.43 \\
    \multicolumn{1}{c|}{\textbf{Water}} & \multicolumn{1}{c|}{\textbf{182/143}} & 97.20 & 95.10 & 79.02 & 95.80 & 95.10 & 95.80 & 91.61 & 100   & 95.8  & 98.60 & 95.8 \\
    \multicolumn{1}{c|}{\textbf{Residential}} & \multicolumn{1}{c|}{\textbf{196/1072}} & 82.09 & 89.09 & 86.19 & 73.41 & 90.95 & 78.54 & 87.59 & 91.11 & 87.87 & 90.39 & 86.01 \\
    \multicolumn{1}{c|}{\textbf{Commercial}} & \multicolumn{1}{c|}{\textbf{191/1053}} & 40.65 & 45.87 & 65.81 & 85.28 & 90.98 & 86.61 & 84.14 & 92.51 & 95.25 & 95.73 & 93.54 \\
    \multicolumn{1}{c|}{\textbf{Road}} & \multicolumn{1}{c|}{\textbf{193/1059}} & 69.78 & 82.53 & 72.11 & 93.95 & 90.46 & 87.72 & 91.78 & 86.87 & 89.71 & 98.21 & 97.07 \\
    \multicolumn{1}{c|}{\textbf{Highway}} & \multicolumn{1}{c|}{\textbf{191/1036}} & 57.63 & 83.20 & 55.21 & 67.08 & 60.91 & 68.82 & 86.20 & 94.66 & 81.18 & 63.42 & 68.53 \\
    \multicolumn{1}{c|}{\textbf{Railway}} & \multicolumn{1}{c|}{\textbf{181/1054}} & 76.09 & 83.87 & 85.01 & 90.89 & 94.46 & 90.23 & 98.58 & 90.56 & 86.34 & 90.70 & 98.86 \\
    \multicolumn{1}{c|}{\textbf{Parking Lot 1}} & \multicolumn{1}{c|}{\textbf{192/1041}} & 49.38 & 70.99 & 60.23 & 88.56 & 99.14 & 98.08 & 92.32 & 90.74 & 92.7  & 91.07 & 100.00 \\
    \multicolumn{1}{c|}{\textbf{Parking Lot 2}} & \multicolumn{1}{c|}{\textbf{184/285}} & 61.40 & 70.53 & 75.09 & 76.14 & 65.26 & 80.35 & 76.84 & 89.92 & 87.02 & 76.49 & 74.74 \\
    \multicolumn{1}{c|}{\textbf{Tennis Court}} & \multicolumn{1}{c|}{\textbf{181/247}} & 99.60 & 100.00 & 83.00 & 100.00 & 100   & 100   & 99.60 & 98.58 & 99.19 & 100.00 & 100.00 \\
    \multicolumn{1}{c|}{\textbf{Running Track}} & \multicolumn{1}{c|}{\textbf{187/473}} & 97.67 & 97.46 & 52.64 & 99.78 & 99.15 & 100   & 98.73 & 98.14 & 89.64 & 99.15 & 100.00 \\ \hline
    \multicolumn{1}{c|}{\textbf{AA}} & \multicolumn{1}{c|}{\textbf{--}} & 80.34 & 86.34 & 77.19 & 88.54 & 91.24 & 91.02 & 90.65 & NA    & 91.82 & 91.95 & 92.45 \\
    \multicolumn{1}{c|}{\textbf{OA}} & \multicolumn{1}{c|}{\textbf{--}} & 77.47 & 84.69 & 78.35 & 86.98 & 91.28 & 89.93 & 91.11 & 92.45 & 91.75 & 91.3  & 92.68 \\
    \multicolumn{1}{c|}{\textbf{\textit{K}}} & \multicolumn{1}{c|}{\textbf{--}} & 0.7563 & 0.8340 & 76.46 & 0.8592 & 0.903 & 0.8910 & 0.8985 & NA    & 0.9033 & 0.9056 & 0.9181 \\
    \bottomrule
    \end{tabular}%
  \label{tab:HSILiDAR}%
\end{table*}%

\subsection{Houston University}

\begin{figure}
  \centering
 \centering
\includegraphics[width=0.95\linewidth]{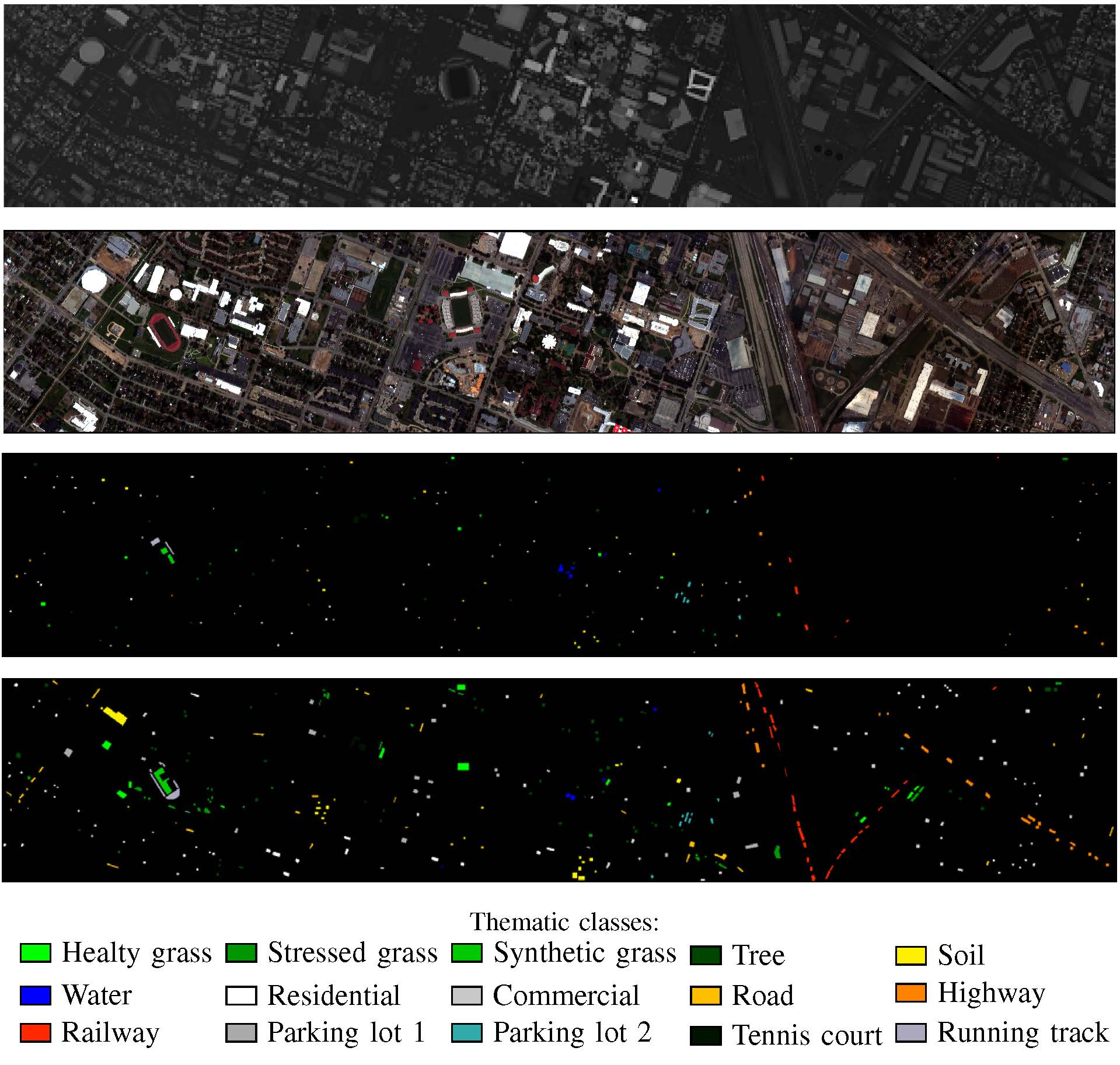}
  \caption{Houston - From top to bottom: LiDAR-derived rasterized DSM, a color composite illustration of the CSI Houston HSI using bands 64, 43, and 22 as R, G, and B, respectively; Training samples; Test samples; and legend of different classes.}
  \label{fig:TTHouston}
\end{figure}

The Houston University data for this section are composed of a LiDAR-derived digital surface model (DSM) and an HSI both captured over the University of Houston campus and the neighboring urban area. This dataset was initially made publicly available for the 2013 GRSS data fusion contest. The HSI and LiDAR data were captured on June 23, 2012 and June 22, 2012, respectively. The size of the dataset is 349 $\times$ 1905 pixels with a ground sampling distance of 2.5 m. The HSI consists of 144 spectral bands ranging 0.38-1.05$\mu m$. Fig. \ref{fig:TTHouston} illustrates the investigated data and the corresponding training and test samples.  The number of training and test samples for different classes are detailed in Table~\ref{tab:HSILiDAR}. 

\subsection{Filtering}

Filtering approaches have been used intensively in the literature to effectively extract contextual and spatial features by attenuating redundant spatial details (based on a criterion) and preserving the geometrical characteristics of the other regions. Among those approaches, one can refer to morphological profiles (MPs \cite{Benediktsson_MP}, i.e., which can be produced by the sequential implementation of opening and closing operators by reconstruction by considering a structuring element of increasing size), attribute profiles (APs \cite{AP_DallaMura}, i.e., which can obtain a multilevel characterization of the input image by considering the repeated implementation of morphological attribute filters), and extinction profiles (EPs \cite{EP-Ghamisi-TGRS-2016}, i.e., which can obtain a multilevel characterization of the input image by considering the repeated implementation of a morphological extinction filter).

These approaches have been investigated frequently for the fusion of LiDAR and HSI since they are fast and conceptually simple and able to provide accurate classification results. For instance, in \cite{Mattia,Ghamisi-IJIDF}, the spatial features of HSI and LiDAR were extracted using APs. Then, they were concatenated and fed to a classifier leading to precise results in terms of classification accuracy in a fast manner. In \cite{HL-CNN-JSTARS-2017}, EPs were used to automatically extract the spatial and elevation features of HSI and LiDAR data. The extracted features were stacked and then classified using a random forest (RF) classifier (i.e., the results obtained by that approach can be found in Table~\ref{tab:HSILiDAR} as $\textbf{EP}_{\textnormal{HSI+LiDAR}}$).   

Filtering approaches such as MPs, APs, and EPs suffer from two shortcomings: The curse of dimensionality and intensive processing time for the subsequent classification steps since they usually increase the number of dimensions by stacking spectral, spatial, and elevation features extracted from HSI and LiDAR, while the number of training samples remains the same. To address this shortcoming, composite kernel- and low rank-based approaches, which will be discussed in the following subsections, have been suggested in the literature to effectively fuse HSI and LiDAR.

\begin{figure}
  \centering
 \centering
\includegraphics[width=0.999\linewidth]{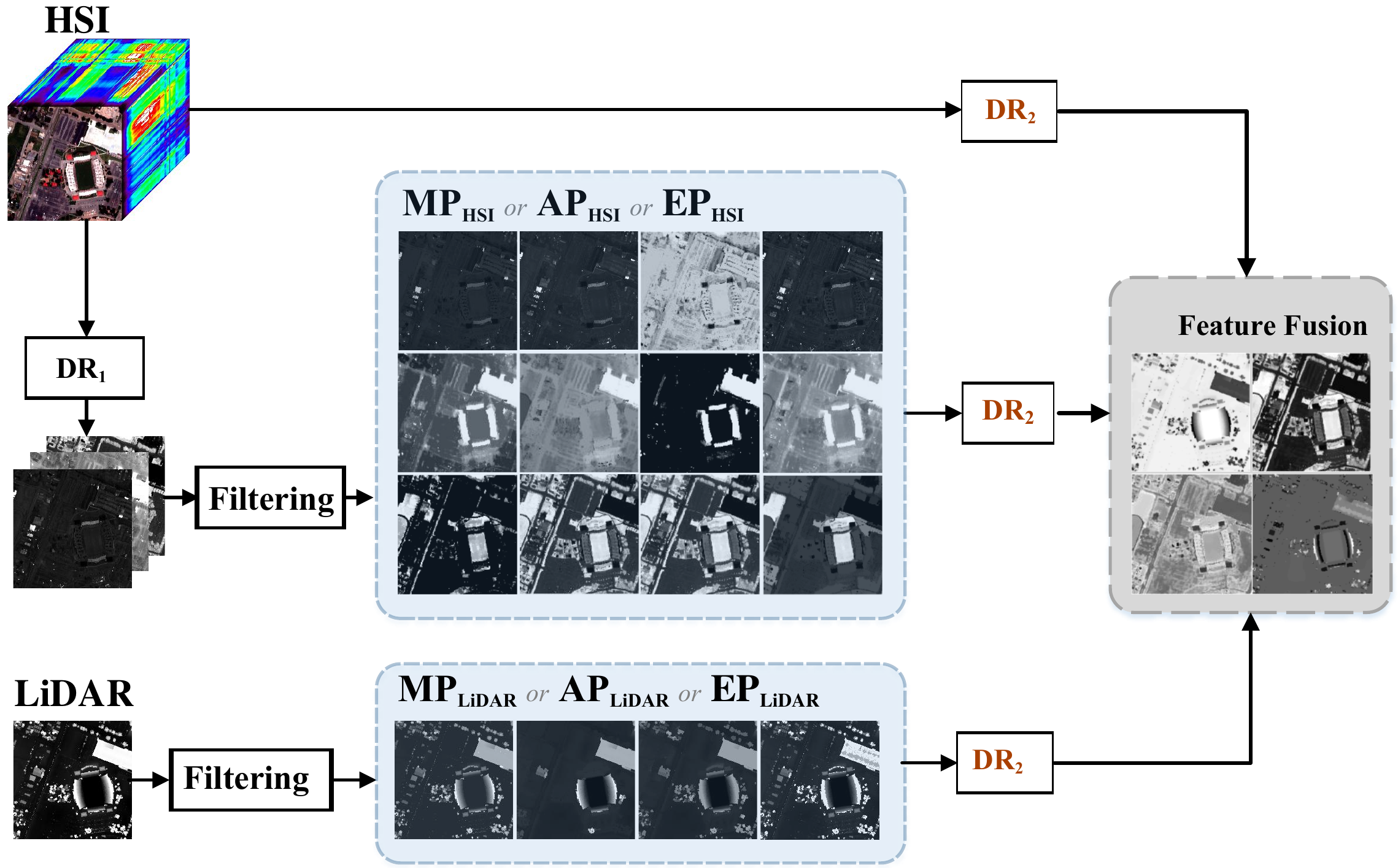}
  \caption{Low-rank models. The use of $\textnormal{DR}_2$ is optional. However, the studies investigated in \cite{HLcomp2013}, \cite{HL-CNN-JSTARS-2017}, \cite{OTVCAfusion} and \cite{SLRCAfusion} recommend the consideration of this extra extra step in order to provide more accurate classification maps.}
  \label{lowrank}
\end{figure}

\subsection{Low-rank models}

To avoid the curse of dimensionality and also increase the efficiency of the analysis compared to filtering approaches, low-rank models were investigated in \cite{HLcomp2013,HL-CNN-JSTARS-2017,OTVCAfusion,SLRCAfusion} whose main assumption was that the extracted features from HSI and LiDAR can be represented into a space of a lower dimension. All those approaches followed a general framework as demonstrated in Fig. \ref{lowrank}. This framework is composed of the following building blocks: 

\begin{enumerate}
\item $\textnormal{DR}_1$ generates base images to build up MP, AP, or EP.
\item Filtering investigates MP, AP, or EP to extract spatial features (e.g., $\textbf{EP/AP/MP}_{\textnormal{HSI}}$) and elevation features (e.g., $\textbf{EP/AP/MP}_{\textnormal{LiDAR}}$) from HSI and LiDAR, respectively.
\item $\textnormal{DR}_2$ is used to produce exactly the same number of spectral, spatial, and elevation features to put the same weight on each category. The other advantages of $\textnormal{DR}_2$ are that it can reduce the executable computational cost as well as noise throughout the feature space. In \cite{HLcomp2013,HL-CNN-JSTARS-2017,OTVCAfusion,SLRCAfusion}, kernel PCA (KPCA) has been use for $\textnormal{DR}_2$.
\item Finally, the outputs of (3) are fused and fed to a classification method. Below, we discuss \cite{HLcomp2013}, \cite{HL-CNN-JSTARS-2017}, \cite{OTVCAfusion} and \cite{SLRCAfusion} in more detail:
\end{enumerate} 

In \cite{HLcomp2013}, the spectral (\textbf{HSI}), spatial ($\textbf{MP}_{\textnormal{HSI}}$), and elevation features ($\textbf{MP}_{\textnormal{LiDAR}}$) were used (as the filtering step). A graph-based feature fusion (GBFF) technique was utilized (as the feature fusion step). Finally, an SVM classifier was used to classify the fused features (results can be found in Table~\ref{tab:HSILiDAR} as \textbf{GBFF}). 

In \cite{HL-CNN-JSTARS-2017}, the spectral (\textbf{HSI}), spatial ($\textbf{EP}_{\textnormal{HSI}}$), and elevation features ($\textbf{EP}_{\textnormal{LiDAR}}$) were concatenated and fed to the GBFF and classified by a 2D CNN. These results can be found in Table~\ref{tab:HSILiDAR} as \textbf{CNNGBFF}.

In \cite{OTVCAfusion}, the following low-rank model was suggested to fuse \textbf{HSI}, $\textbf{EP}_{\textnormal{HSI}}$, and $\textbf{EP}_{\textnormal{LiDAR}}$:  
\begin{equation}\label{eq: D2D1model1}
	{\bf F} = {\bf A}{\bf V}^T + {\bf N},
\end{equation}
where ${\bf F}=\left[{\bf f}_{\left(i\right)}\right]$ is an $n \times p$ matrix which contains the vectorized features in its columns, ${\bf V}$ is a $p\times r$ unknown matrix containing the subspace basis, ${\bf A}=\left[{\bf a}_{\left(i\right)}\right]$ is a $n\times r$ matrix which contains the $r$ unknown fused features in its columns, and ${\bf N}=\left[{\bf n}_{\left(i\right)}\right]$ is the model error and noise. Note that $r$ is the number of fused features. Also, hyperspectral bands, and hyperspectral and LiDAR features are concatenated in matrix ${\bf F}$ (${\bf F}=\left[\textbf{EP}_{\textnormal{HSI}}, {\bf HSI}, \textbf{EP}_{\textnormal{LiDAR}}\right]$).

In model (\ref{eq: D2D1model1}), matrices ${\bf A}$ and ${\bf V}$ are both unknown. Therefore, they both need to be estimated. In \cite{OTVCAfusion}, orthogonal total variation component analysis (OTVCA) \cite{OTVCA} was suggested to solve this problem (as the feature fusion step shown in Fig.~\ref{lowrank}). OTVCA is given by 
\begin{equation}\label{eq: cost}
	\arg\min_{{\bf A},{\bf V}}~\frac{1}{2}\left\|{\bf F}-{\bf A}{\bf V}^T\right\|^{2}_{F}+\lambda\sum_{i=1}^r\mbox{TV}({\bf a}_{(i)})~\mbox{s.t.}~ {\bf V}^T{\bf V}={\bf I}_r,
\end{equation}
where the total variation penalty (TV) is applied spatially on the fused features. TV preserves the spatial structure of the features while promotes piece-wise smoothness on the fused features. As a result, the final classification map contains homogeneous regions. The OTVCA fusion results can be found in Table~\ref{tab:HSILiDAR} as \textbf{OTVCA}.

In \cite{SLRCAfusion}, the extracted features were defined using the sparse and low-rank model given in \cite{WSRRR}, 
\begin{equation}\label{eq: D2D1model}
	{\bf F} = {\bf DW}{\bf V}^T + {\bf N},
\end{equation}
where ${\bf D}$ is an $n\times n$ matrix which contains two-dimensional wavelet basis, and ${\bf W}=\left[{\bf w}_{\left(i\right)}\right]$ is an $n \times r$ matrix containing the unknown 2D wavelet coefficients for the $i$th fused component. In \cite{SLRCAfusion}, the sparse and low-rank component analysis \cite{WSRRR,SLRCA} was used to estimate ${\bf W}$ and {\bf V} given by
\begin{equation}\label{eq: cost1}
	\arg\min_{{\bf W},{\bf V}}~\frac{1}{2}\left\|{\bf F}-{\bf DW}{\bf V}^T\right\|^{2}_{F}+\lambda\sum_{i=1}^r\left\|{\bf w}_{(i)}\right\|_1 ~\mbox{s.t.}~ {\bf V}^T{\bf V}={\bf I}_r,
\end{equation}
Note that the estimated fused features are given by ${\hat{\bf F}_{fused}}= {\bf D}{\hat{\bf W}}$. The fused features are expected to be sparse in the 2D wavelet basis. Therefore, in \cite{WSRRR}, to enforce the sparsity an $\ell_1$ penalty on the wavelet coefficients ${\bf W}_r$ was used. As a result, promoting sparsity on the fused feature improves the SNR and the final classification accuracies. Results for this approach can be found in Table~\ref{tab:HSILiDAR} as \textbf{SLRCA}.


\subsection{Composite kernel}

Composite kernel-based fusion approaches partially overcome the shortcomings of the filtering approaches by designing several kernels to handle spectral, spatial, and elevation features in feature space \cite{CampsValls2006}. 

In \cite{Zhang-RSL-2017}, spectral, spatial (e.g., $\textbf{EP}_{\textnormal{HSI}}$), and elevation (e.g., $\textbf{EP}_{\textnormal{LiDAR}}$) information were fused using a local-region filter (LRF) and composite kernels. Results for this approach can be found in Table~\ref{tab:HSILiDAR} as \textbf{FFCK}. The main shortcoming of this approach was that its obtained classification accuracy was dramatically influenced by the $\mu$ parameter which represents the amount of trade-off between the spectral and spatial-elevation kernels. To solve this issue, in \cite{MkhodadadzadehLH}, a fusion approach was introduced capable of exhibiting substantial flexibility to integrate different feature sets without requiring any regularization parameters. That approach was based on APs and multiple feature learning using the subspace multinomial logistic regression (MLR\textit{sub}) classifier. The result of this approach is shown as \textbf{MLR\textit{sub}} in Table~\ref{tab:HSILiDAR}.  

A joint sparse representation classification approach was proposed in \cite{JSRC_2016} for multisource data fusion where the multisource data were weighted to have better sparse representation. The core idea of this approach was based on sparse representation classification. Then, the minimum distance (in the sense of $\ell_2$ norm) between each sample and its sparse representation using subdictionaries (containing only training samples for one class) were used to allocate the class labels. However, in \cite{JSRC_2016} the regularization term was weighted according to the data sources. Moreover, the method was also translated into the kernel space using kernel tricks. Results for the composite kernel version can be found in Table~\ref{tab:HSILiDAR} as \textbf{ALWMJ-KSRC}.

\subsection{Deep learning}

Hyperspectral imaging often exhibits a nonlinear relation between the captured spectral information and the corresponding material. This nonlinear relation is the result of several factors such as undesired scattering from other objects in the acquisition process, different atmospheric and geometric distortions, and intraclass variability of similar objects. This nonlinear characteristic is further magnified when we deal with multisensor data. On the other hand, deep architectures are inherently able to extract high-level, hierarchical, and abstract features, which are usually invariant to the nonlinearities of the input data.

\begin{figure*}
  \centering
 \centering
\includegraphics[width=0.99\linewidth]{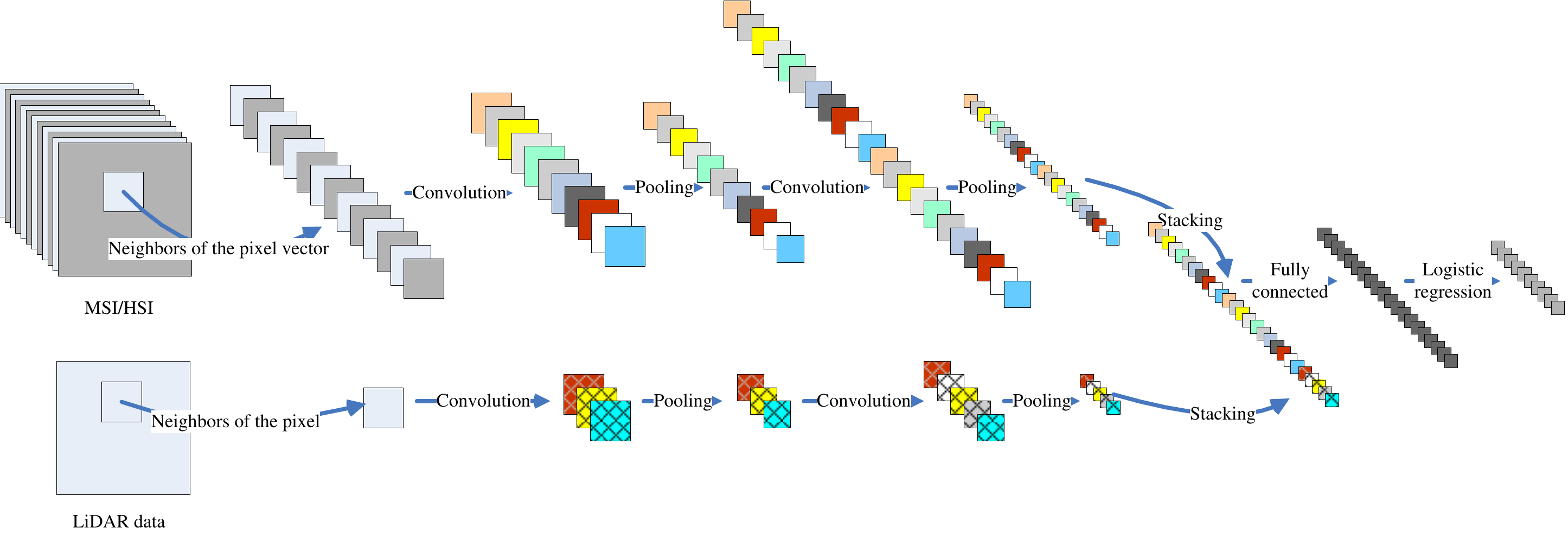}
  \caption{Deep learning models \cite{Chen-LH-GRSL-2017}.}
  \label{DL}
\end{figure*}

Deep learning is a fast-growing topic in the remote sensing community whose trace can also be found in the research area of LiDAR and HSI data fusion. For instance, in \cite{HL-CNN-JSTARS-2017}, a classification method was developed to fuse spectral (\textbf{HSI}), spatial ($\textbf{EP}_{\textnormal{HSI}}$), and elevation features $\textbf{EP}_{\textnormal{LiDAR}}$ using a 2D-CNN and GBFF. The results for this approach can be found in Table~\ref{tab:HSILiDAR} as \textbf{CNNGBFF}. To extract spatial and elevation features in a more effective manner than in \cite{HL-CNN-JSTARS-2017}, two distinct CNN streams (as shown in Fig.~\ref{DL}) were employed in \cite{Chen-LH-GRSL-2017}. The heterogeneous features obtained by the previous CNNs were then classified by a fully connected deep neural network. In \cite{rs10101649}, a three-stream CNN with multisensor composite kernel was utilized to fuse spectral, spatial, and elevation features.

\subsection{Trends of hyperspectral and LiDAR fusion}

The following trends for the advancements of hyperspectral and LiDAR fusion need to be further investigated in the future:
\begin{itemize}
\item Due to the increased availability of large-scale DSMs and hyperspectral data, the further development of fast, accurate, and automatic classification/fusion techniques for the challenging task of transferable and large-area land-cover mapping is of great interest.
\item Investigation of the advanced machine learning approaches (e.g., deep learning, domain adaptation, and transfer learning) for developing transferable classification/fusion schemes of areas with limited number of training samples is in demand in our community. 
\item The development of sparse, low-rank, and subspace fusion approaches is another interesting line of research to address the high dimensionality of the heterogeneous features extracted from HSI and LiDAR to further increase the quality of classification outputs. 
\item \cite{cGAN-GhamisiYokoya} took the first step in the remote sensing community to simulate DSM from single optical imagery. This work opens a new path in front of researchers to further modify this approach and design more sophisticated network architectures to produce more accurate elevation information from single optical images.
\end{itemize}

As stated above, the classification/fusion of large-scale data (e.g., big data) is a vitally important research line which will be further discussed in the next section.

\section{Multitemporal Data Fusion}
\begin{figure*}
\begin{center}
\includegraphics[width=0.999\linewidth]{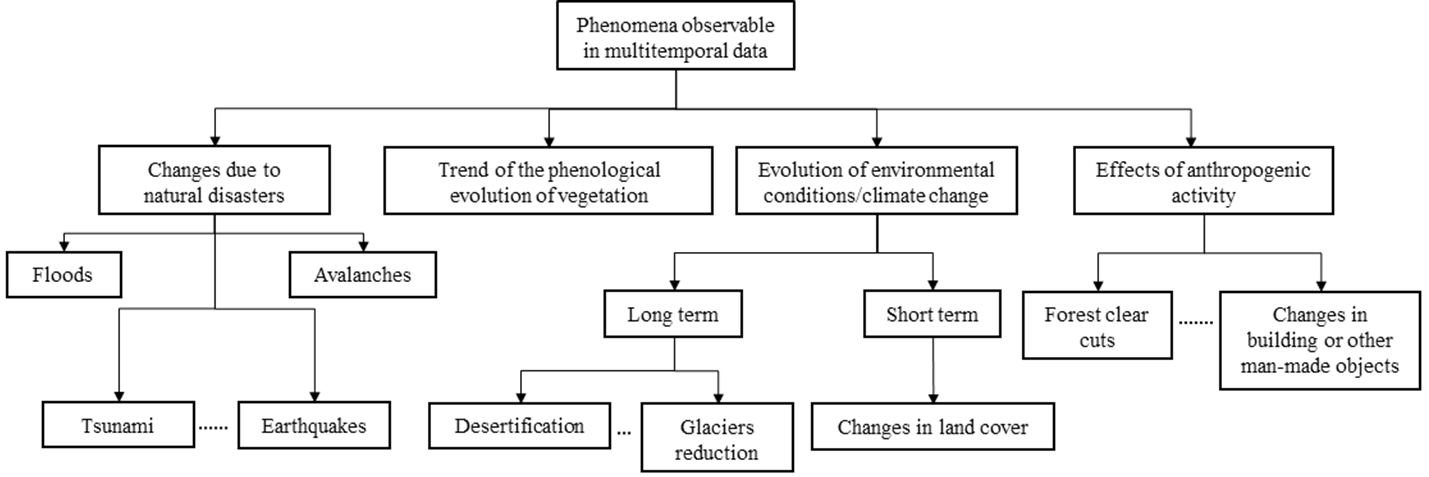}
\end{center}
\caption{Example of taxonomy of phenomena observable and detectable by using temporal information in remote sensing data analysis.}
\label{fig:F1MDF}
\end{figure*}

The use of multitemporal information is crucial for many important applications (from the analysis of slow and smooth evolving phenomena \cite{J24,J25} to steep and abrupt changes \cite{J14,J22,J26}). Fig. \ref{fig:F1MDF} shows a taxonomy of temporal phenomena that can be observed and detected by including the time variable in the analysis of remote sensing data. 

As discussed earlier in the introduction part, the recent availability of satellite constellations like Sentinel 1 and Sentinel 2 \cite{J27}, which are characterized by the acquisition of fine resolution images (up to 10 m) with a very short revisit time (few days depending on the latitude), is making the time dimension of satellite remote sensing images one of the most important sources of information to be exploited for the extraction of semantic content from a scene \cite{J14}. The time variable can expand the dimensionality of interest from 3D to 4D in space and time and can be exploited working with pairs of images, short time-series or long time-series of either multispectral passive or SAR active images \cite{J14,J16}. Moreover, it is also possible to fuse together time-series of multisensor images in a proper multitemporal sensor-fusion framework \cite{J28,J29,J30}. Fusion of temporal information with spatial and/or spectral/backscattering information of the images opens the possibility to change the perspective also from the viewpoint of methodologies for data analysis. We can move from a representation of 3D cubes with multispectral images to 4D data structures, where the time variable adds new information as well as challenges for the information extraction algorithms.

Analyzing the literature, the most widely addressed applications of multitemporal data are the analysis/classification of image time-series and change detection \cite{J16,J14}. Nonetheless, there are many emerging topics based on the joint exploitation of the spectral, spatial, and temporal information of long and dense time-series of fine spatial resolution images \cite{J21,J31,J32,J33}. These topics were investigated widely in the past with coarse/medium spatial resolution images (i.e., at the scale of MODIS or MERIS/ENVISAT). However, with the availability of relatively dense time-series of fine resolution images (e.g., Sentinel 1, Sentinel 2, and Landsat-8), it is now possible to develop studies at a dramatically increased resolution. For example, it is possible to study the phenology of the vegetation in specific local areas or to analyze the trends of vegetation in single fields of agricultural regions for precision farming applications \cite{J21,J25}. This can be done also by fusing the acquisitions of different satellites in a single multisensor time-series. Given the complexity and the extent of the topic, in the following we analyze temporal information in relation to the classification problem, which is one of the most challenging lines of research and widely studied in the past \cite{J25,J31,J34}. First, we analyze the definition of the different classification problems with multitemporal data and briefly recall the methods presented in the literature for the solution of these problems. Then, we discuss the challenges related to multitemporal classification especially from the view point of the availability of labeled training data.

\subsection{Multitemporal Information in Classification}

\begin{figure*}
\centering
\includegraphics[width=0.999\linewidth]{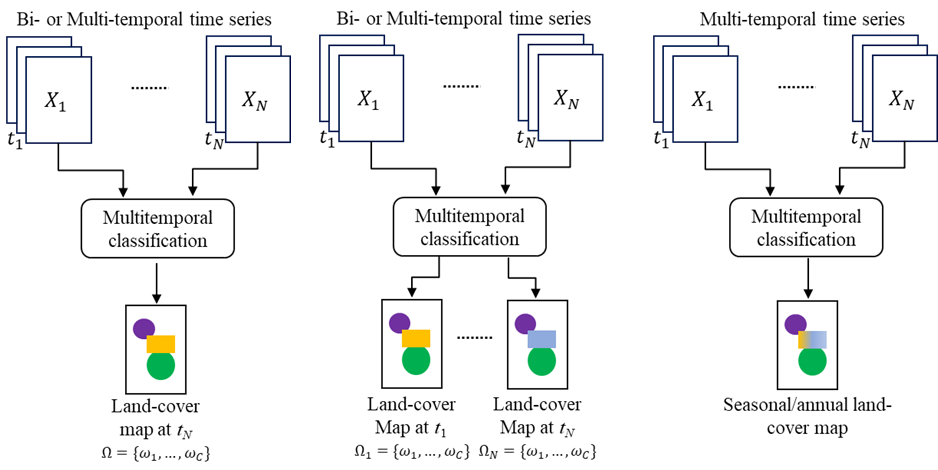}
\\
(a)~~~~~~~~~~~~~~~~~~~~~~~~~~~~~~~~~~~~~~~~~~~~~~~(b)~~~~~~~~~~~~~~~~~~~~~~~~~~~~~~~~~~~~~~~~~~~~~~~~~~~(c)\\
\hspace{1cm}
\caption{Block scheme for achieving different goals in multitemporal classification: (a) a land-cover map associated with the most recent image of a time-series; (b) a land cover map for each item of the time-series; (c) an annual/seasonal land-cover map with classes that represent the behavior of the temporal signature of each pixel/region in the time-series.}
\label{fig:F2MDF}
\end{figure*}

Let us assume the availability of a set of multitemporal images (a time-series with many images or at least a pair of images) acquired of the same geographical area at different times. The classification of these multitemporal data can be defined in different ways depending on the objective of data analysis. The goal of the classification can be to generate: i) a land-cover map associated with the most recent image (acquisition) of a time-series (or of a pair of acquisitions) (Fig. \ref{fig:F2MDF} (a)) \cite{J17}; ii) a land cover map for each item of the time-series, thus, producing a set of multitemporal land-cover maps (Fig. \ref{fig:F2MDF} (b)) that also implicitly models the land-cover transitions \cite{J18,J35}; iii) an annual/sesonal land-cover map with classes that represent the behavior of the temporal signature of each pixel/region in the images in a year/season (Fig. \ref{fig:F2MDF} (c)) \cite{J19,J5}. These three definitions should result in significantly different classification approaches based on different assumptions. Unfortunately, in many cases in the literature and in the definition of application-oriented systems, the problems are not properly identified and modeled with the implication of obtaining sub-optimal results.

The use of the temporal information in classification dates to the early 1980s. First approaches used a stacked vector representation of the multitemporal data as input to the classifiers resulting in the so-called supervised direct multidate classification \cite{J22}. The main idea of such approaches is to characterize pixels by stacking the feature vectors of the images acquired at two (or more) times. Then the classification is carried out by training the classifiers to produce a map describing only the land covers of the most recent image. However, this is theoretically affordable under the assumption that both: i) there are no changes in the land covers between the considered image acquisition dates; and ii) it is possible to properly model the complexity of the data distributions with respect to the classification methodology. The latter becomes critical when statistical Bayesian approaches are used. Another possible way of using multidate direct classification approaches is to classify the land covers of each item of the time-series, thus, producing a land-cover map for each available acquisition time. This allows one to explicitly identify land-cover transitions and to remove the assumption that there are no changes between the considered dates. However, a proper modeling of the change information requires the availability of labeled training data that can adequately represent the statistics of all possible combination of classes, including those associated to the changes. This is seldom possible in real application scenarios. 

Many methodologies have been developed to address the above-mentioned issues of multidate direct classification. Swain \cite{J23} introduced in a pioneering paper a very interesting approach to the classification of multitemporal data based on the cascade classification of pairs of images. He modeled the classification problem from a Bayesian perspective introducing the temporal correlation in the classification process of images acquired at different times for linking class probabilities estimated on single images. In this way it is possible to decouple in the multitemporal classification problem the modeling of the class distributions at each single date with the estimation of the temporal correlation between images. Bruzzone et al. \cite{J17} developed and generalized this framework to the case of multitemporal and multisensor data, introducing an approach to compound classification based on neural networks classifiers being able to properly merge in a Bayesian decision framework the distribution free estimations of the class parameters derived from both multispectral and SAR multitemporal images. This kind of fusion has been studied widely in the past two decades, and developed in the context of different classification methodologies including several neural models (e.g., multilayer perceptron neural networks, radial basis function networks), kernel methods (e.g., support vector machines \cite{J8,J9}), and multiple classifier systems \cite{J3,J4,J5} (e.g., based on the fusion of neural and statistical classification algorithms). Also the joint exploitation of the spatio-temporal information has been investigated including Markov Random Fields in the modeling of the spatio-temporal context of multitemporal data. 

Nowadays, the challenge, still poorly addressed, is to exploit deep learning architectures (e.g., convolutional neural networks) in the classification of multitemporal data \cite{J36,J37}. These architectures are intrinsically able to capture the spatio-temporal patterns in the definition of the classification model and, thus, to increase the accuracy of the land-cover/land-cover-transition maps. However, there is still a very significant challenge to define theoretically sound and computational affordable deep learning architectures able to properly process multitemporal images. Indeed, the use of the 4D data structure sharply increases the complexity associated with deep learning architectures and requires an amount of training data that currently is far from being available.

\subsection{Challenges in Multitemporal Classification}

The main challenges associated with the exploitation of the time information source in the classification of remote sensing data are related to the availability of adequate labeled samples for the definition of training sets suitable for the learning of supervised algorithms. The problem is to define statistically significant training sets able to represent the structured information content present in the data \cite{J1,J18}. This problem is much more critical than in the classification of single images given the complexity associated with the possible combinations of land-cover classes in the spatio-temporal domain. A proper modeling of the temporal information would require multitemporal ground reference samples (or reliably annotated multitemporal images) with samples that represent: i) all the multitemporal classes; ii) the inter-relation between classes along the time-series (e.g., land-cover transitions or different kinds of changes) with a reliable statistic; and iii) the high temporal and spatial variabilities in large scenes. These constraints are very difficult to satisfy in real applications. For this reason, a large attention has been and is still devoted to the use of methods that address the limitations of the real training set. 

In this context, the scientific community activities have been focused on semi-supervised (also called partially unsupervised) classification methods \cite{J1,J2,J10,J11,J12,J13}. These methods jointly exploit the available labeled training data and the distribution of the observed images for improving the modeling of the spatio-temporal properties of the analyzed time-series. Early attempts to use these approaches in remote sensing are related to the use of the expectation-maximization algorithm in the context of land-cover map updating with a maximum likelihood classifier \cite{J1}. This has been extended to the use with the cascade and compound classification approaches to the classification of bi-temporal images \cite{J2,J17}. The approaches can integrate multispectral and SAR multitemporal data, as well as multisensor images. The problem of semi-supervised learning with multitemporal data has been then formulated in the more general theoretical problem of domain adaptation for which different solutions can be found in the literature \cite{J20}. For example, the use of semi-supervised SVM has been widely investigated with different methodological implementations \cite{J8}. The use of active learning in the framework of compound classification for optimizing the definition of training data while minimizing the cost associated with their collection was proposed in \cite{J10}. The main idea was to collect ad-hoc training samples in portions of the images where there is high multitemporal uncertainty on the labels of the classes. Transfer learning approaches were proposed in \cite{J11,J12,J13}, where change detection-based techniques were defined for propagating the labels of available data for a given image to the training sets of other images in the time-series. The main observation at the basis of these techniques is that the available class labels can be propagated within the time-series to all the pixels that have not been changed between the considered acquisitions. In this way, unsupervised change detection can become a way to increase the amount of supervision that can be injected in the learning of a multitemporal classifier.

However, despite the large amount of papers on methods capable to capture in a semi-supervised way the information of 4D data structures, this is still a critical issue in the classification of multitemporal data and definitely an open issue in the framework of multitemporal classification with deep learning architectures. In this last case the challenges are related to decouple in the architecture of the network the learning of the relation between the spatio-temporal patterns to achieve feasible requirements on the amount and the characteristics of training samples without degrading significantly the capability to extract the semantic of spatio-temporal pattern from the data. We expect that these crucial issues will be widely addressed in future years.

\section{Big Data and Social Media}
\label{Sect:bdsm}

In the recent decade, big data has become a very important topic in many research areas, e.g., remote sensing applications~\cite{Chi16PIEEE}. Every day a massive number of remote sensing data is provided by a large number of Earth observation (EO) space borne and airborne sensors from many different countries. In the near future, all-day, all-weather and full spectrum acquisition segment datasets will be provided by commercial satellites, such as the Jilin-1 constellation, which has launched 10 fine-spatial resolution satellites by February 2018 and will have 60 satellites in orbit by 2020 with a capability of observing any global arbitrary point with a 30 minute revisit frequency~\cite{JilinCon18}. Those video satellites with a (very) fine temporal and spatial resolution can effectively be exploited to monitor our location-based living environments like CCD cameras but on a much larger scale~\cite{Chi17PIEEE}. From a broader spatial perspective, new opportunities for humankind can be provided by the big remote sensing data acquired by those satellites jointly with social media data providing local and live/real time information to better monitor our living environment~\cite{Chi17PIEEE}, especially in the applications of smart cities~\cite{Shapiro06res,Gamba13PIEEE}, emergency and environmental hazards~\cite{Eismann09PIEEE,Serpico12PIEEE}, etc.. On the other hand, new challenges can appear from unprecedented
access to a huge number of remote sensing data
that are leading to a data-rich but knowledge-poor
environment in a fast manner. Here, the semantic gap of remote sensing
data is usually caused due to the lack of certain land-cover
or land-use remote sensing categories on site. As an example,
one can analyze the change in remote sensing images before and
after floods. In this context, it is possible to roughly determine the damaged area
by unsupervised learning algorithms, but it is difficult to assess
the details (e.g., the damage to transportation infrastructure) ~\cite{Chi17PIEEE}.

Social media data provides one of the most important data sources from human activities and are usually comprised of geolocated posts, tweets, photos, video and audio with rich spatial information. With the fast development of computer technologies and internet innovations, social media data are easily created by wearable and intelligent mobile devices equipped with Global Position System (GPS) receivers. Those data can be disseminated quickly to social networks like Facebook, Twitter, YouTube, Weibo and in particular, messaging apps like SnapChat, WhatsApp, and WeChat. Accordingly, the big data produced by the integration of massive global-view remote sensing data and local-but-live location-based social media data can offer new opportunities for smart city and smart environment applications with the ``ground reference" collection through social sensing~\cite{SocialSensing13Aggarwal} and crowd sensing~\cite{Crowdsensing11Ganti}, in particular for hazards and disaster identification or tracking~\cite{Barrington2011Crowdsourcing,schnebele15,Cervone15boulderflood,Chi17PIEEE}.

\begin{figure}
\vskip 0.2in
\begin{center}
\includegraphics[width=0.999\linewidth]{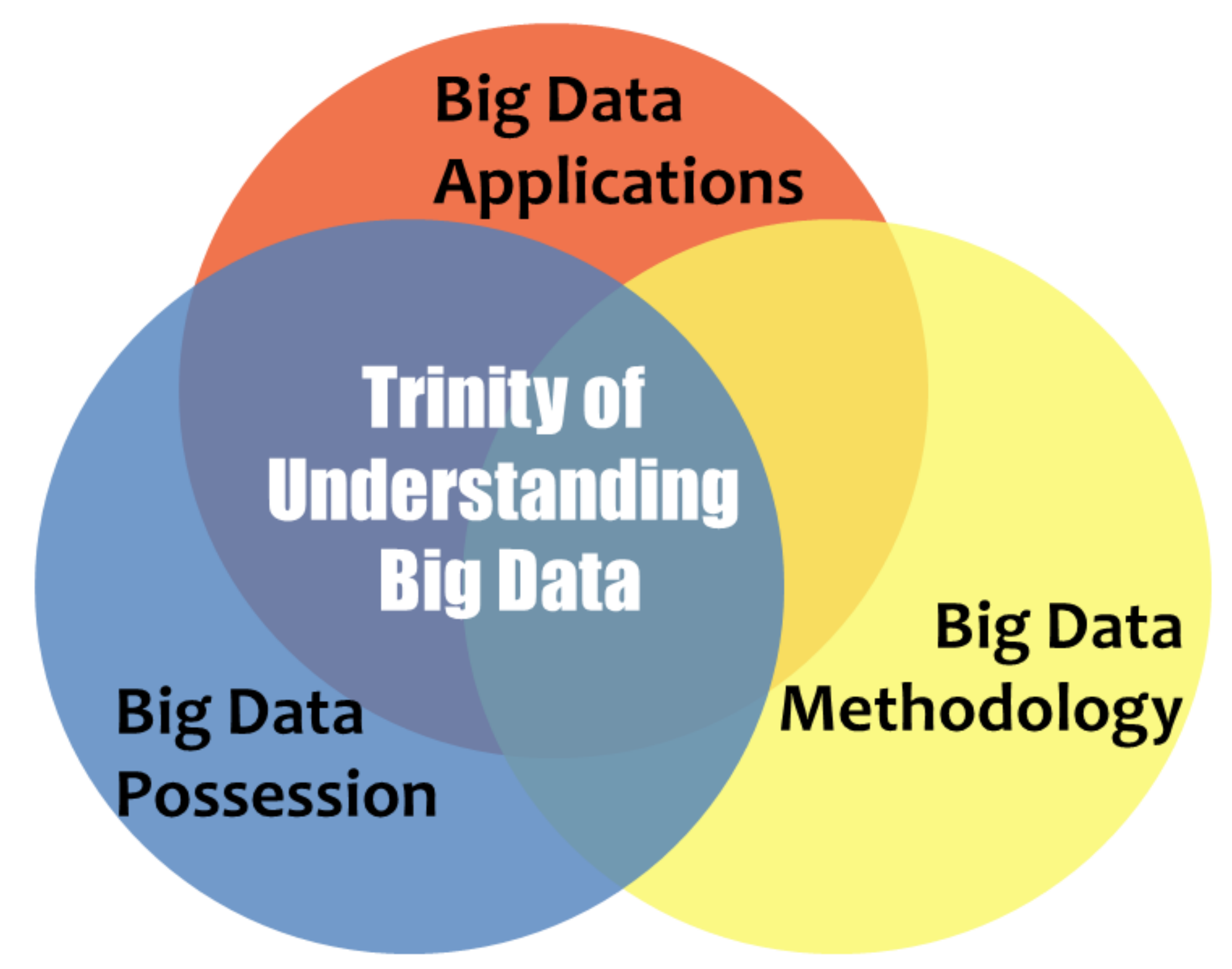}
\end{center}
\caption{Trinity for understanding big data, i.e., three facets of big data from different perspectives related to who owns big data, who has innovative big data methods and methodologies, and who needs big data applications~\cite{Chi16PIEEE}. This list can further be extended by big data visualization and big data accelerated computing.}
\label{fig:trinity}
\end{figure}

To better analyze and utilize big data in remote sensing with social media data, as in the definition of connotations of big data in~\cite{Chi16PIEEE}, it can be expressed in the context of a trinity framework with three perspectives, i.e., owning data, data applications and data methods. The trinity concept of big data is illustrated in Fig.~\ref{fig:trinity}. Accordingly, different perspectives have individual challenges and all the facets of such a trinity share common challenges, which have been discussed in detail in~\cite{Chi16PIEEE}.

To derive the value of big data, combining remote sensing and social medial data, one of the most important challenges is how to process and analyze those data by novel methods or methodologies. Since remote sensing data have significantly different properties from those of social media data, typical data fusion methods cannot be exploited directly for combining remote sensing and social medial data. Often, remote sensing data consist of multi-source (laser, radar, optical, etc.), multi-temporal (collected on different dates), and multi-resolution (different spatial resolution) data. Most remote sensing data are images. Social media data have a much wider array of formats, including images, videos, audio and texts, where texts contain different types of textual information, such as geotagging, hashtags, posts, tweets, RSS (Rich Site Summary), etc.. Nevertheless, the challenges for data fusion of remote sensing and social media data are similar to those in a general big data problem, i.e., data representation, data analysis, and data accelerated computing~\cite{Chi16PIEEE}. However, advances of artificial intelligence (AI) techniques in particular, deep neural networks, have merged data representation and data analysis to a unified AI model. In recent decades, high performance computing has developed for data accelerated-based computing on big data platforms, such as Hadoop~\cite{Hadoop2009White} or SPARK~\cite{Spark2010Zaharia}. In particular, with the fast development of artificial intelligence (AI), GPU-accelerated computing by using a graphics processing unit (GPU) and AI (in particular deep learning) chips have developed quickly in recent years for accelerating deep learning computing in a heterogeneous platform by combining CPU, GPU, FPGA, etc..

\begin{figure*} \begin{center} \begin{tabular}{ccc}
{\includegraphics[width=0.27\linewidth]{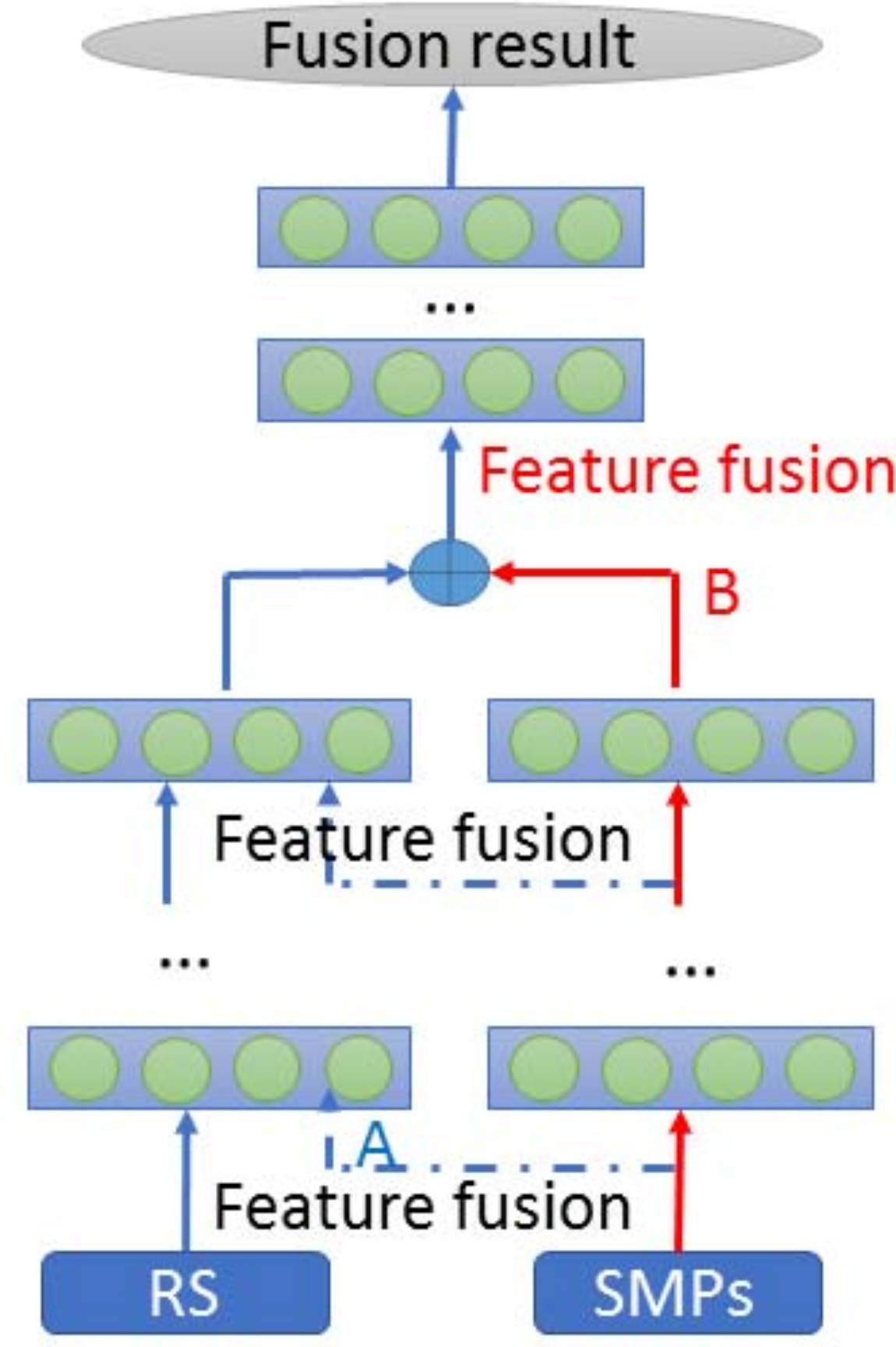}\label{fig:feature}} &
{\includegraphics[width=0.27\linewidth]{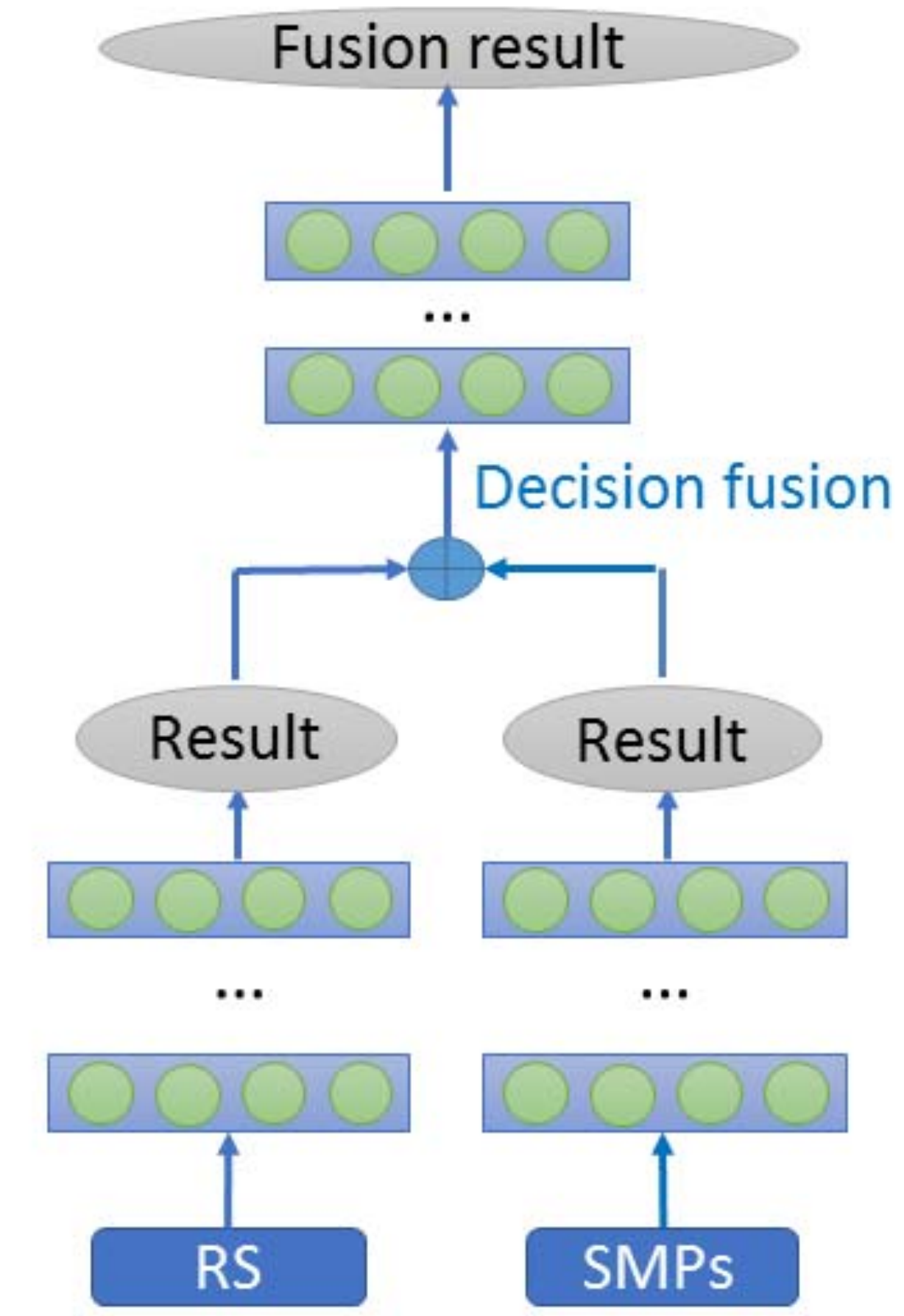}\label{fig:decision}} &
{\includegraphics[width=0.45\linewidth]{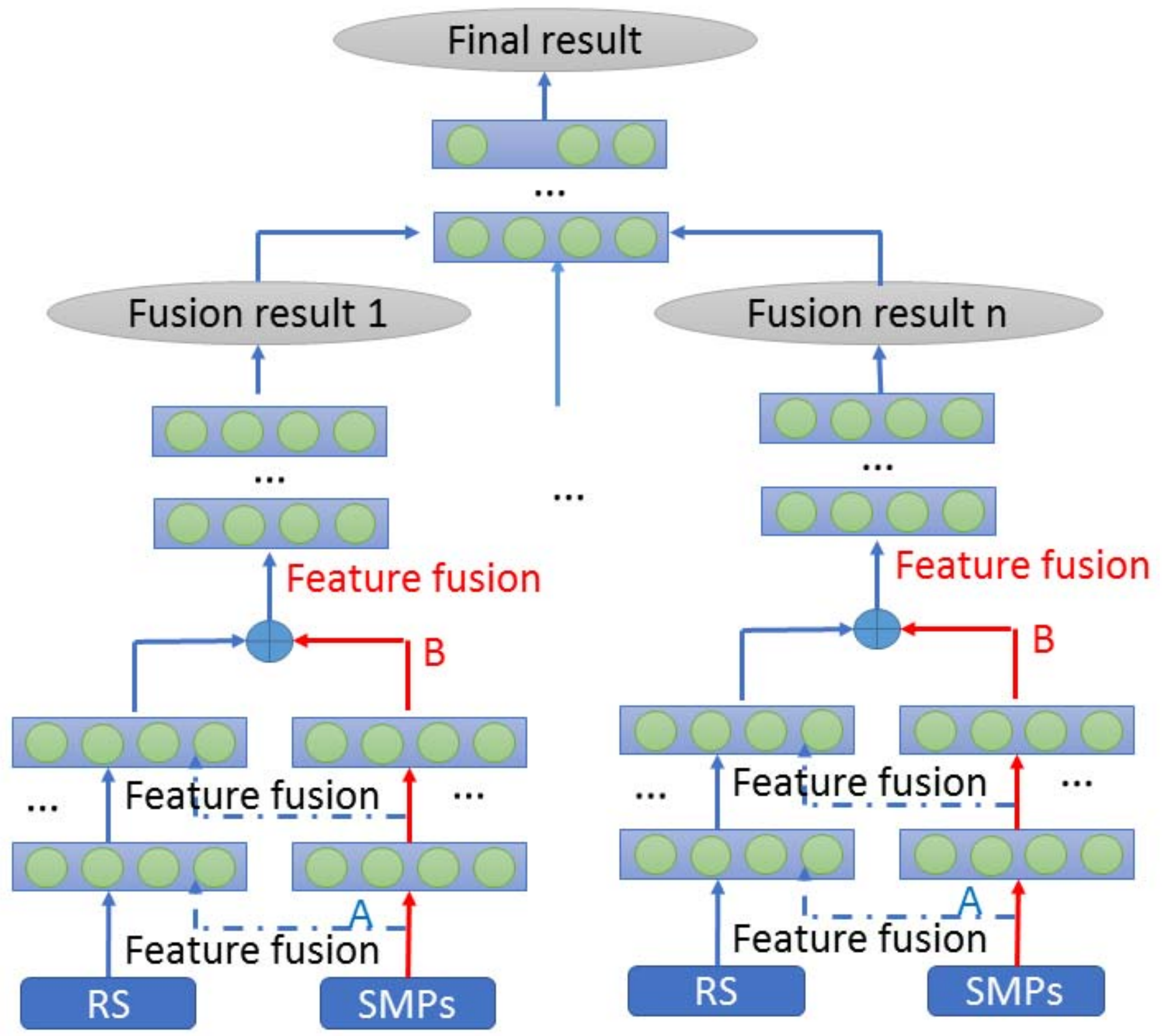}\label{fig:featureDecision}} \\
(a)&(b)&(c)\\
\end{tabular} \end{center} \caption{On the context of deep neural networks, the integration of remote sensing (RS) images with social media photos (SMP): (a) Feature-based fusion, (b) Decision-based fusion, and (c) Feature-decision-based fusion, respectively.}
\label{fig:fusion}
\end{figure*}

Social media data, such as photos with geotaggings, can be integrated with remote sensing data in feature or decision or feature-decision levels, respectively, in the context of deep neural networks. In the feature-based data fusion, social media photos (SMPs) can be integrated to the same deep neural network to extract the features for further processing as shown in Fig.~\ref{fig:fusion}(a). Here, the feature extractor can have a deep architecture and the features generated from SMPs can be integrated in each layer (or arbitrary layers) of the deep neural network. Nonetheless, features can be extracted individually from remote sensing images and the SMPs by different deep neural networks as shown in Fig.~\ref{fig:fusion}(b). After that, the fused features can be sent to a deep neural network for further processing with feature convolutional layers, activation layers, pooling layers, and final classification to get a more reliable and more accurate result. In the decision-based fusion, each deep neural network is designed to firstly extract the features of remote sensing images or social media photos, and then the classification result is generated by individual features. In the decision level, those results provided by the remote sensing and social media data, respectively, are integrated to a unified deep neural network as shown in Fig.~\ref{fig:fusion}. In this case, social media data can have diverse types, such as images, texts and so on, such that different types of the social media data can build different deep neural networks for further decision fusion. By combining two properties of feature-based and decision-based fusion strategies, the feature-decision-based fusion can be easily derived based on deep neural networks. The challenge is how to design a unified DNN model to efficiently and accurately fuse the heterogeneous data.

\begin{figure}
  \centering
 \centering
\includegraphics[width=0.999\linewidth]{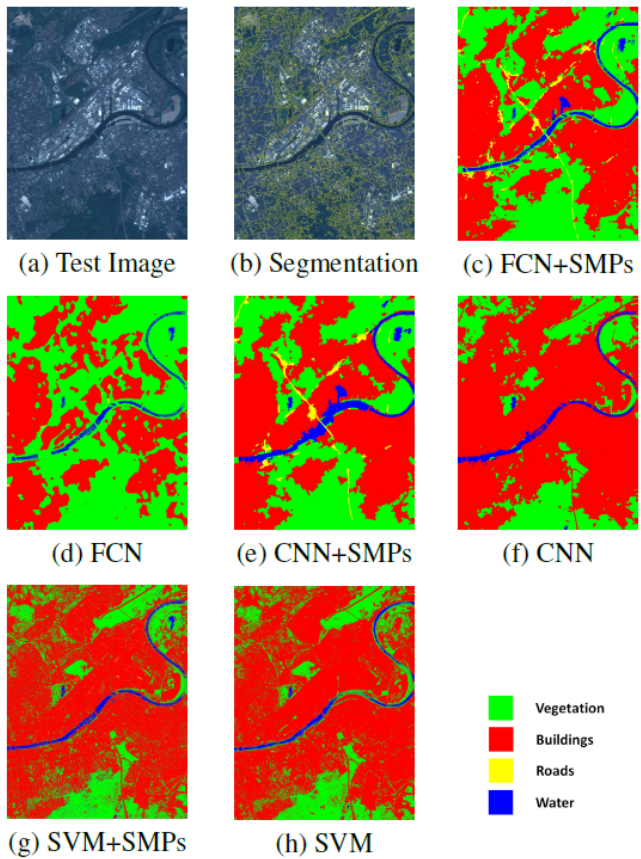}
  \caption{Several classification maps obtained by SVM, CNN, and FCN with or without using SMPs.}
  \label{fig:BIG}
\end{figure}

Except for directly modeling a deep learning algorithm by the integration of remote sensing and social media data, social media photos can be utilized to label remote scene images, especially for fine spatial resolution data. For instance, SMPs with the same positions as the fine resolution remote sensing data can be acquired to help volunteers without any professional knowledge to effectively label remote sensing scene images. To validate the effectiveness of using SMPs for labeling, fine spatial resolution remote sensing images in Frankfurt, Germany acquired by the Jilin-1 satellite are utilized for remote sensing image classification. The classification models are trained respectively on the training datasets labeled with and without SMPs. Table~\ref{tab:tabSM1} shows the prediction results on the test data. Both the fully convolutional network (FCN) model \cite{DBLP:journals/corr/LongSD14} and the CNN model are constructed based on the pre-trained ImageNet VGG-16 network \cite{Simonyan14c} with the cross-entropy loss. The SVM model with the RBF kernels is adopted for a further comparison. Fig. \ref{fig:BIG} illustrates Several classification maps obtained by SVM, CNN, and FCN with or without using SMPs. 
\begin{table}
    \caption{The overall accuracies are compared in terms of the training datasets labeled by volunteers with social media photos (SMPs) and without SMPs, respectively.} \label{tab:tabSM1}    \centering
    \begin{tabular}{ccc}
    \hline
    \multicolumn{2}{c}{Models} & \multicolumn{1}{c}{OA$\%$}\\
       \hline
    \multirow{3}{*}{With SMPs}
    \centering
      & FCN &78.91\\
      & CNN &74.85\\
      & SVM &62.40\\
    \hline
    \multirow{3}{*}{Without SMPs}
    \centering
      & FCN &71.23\\
      & CNN &65.72\\
      & SVM &61.16\\
      \hline
    \end{tabular}
\end{table}

Except for the challenge of designing a novel data technology by fusing big remote sensing data combining remote sensing and social media data, how to exploit the two different types of data is another challenging problem. As future trends, big remote sensing data will be utilized for monitoring natural hazards as well as human-made disasters, such as factory explosion. In addition, big remote sensing data can provide rich information on the contents and locations of the scenes for Augmented Reality (AR) applications, 3D reconstruction, indoor positioning, and so on.
\section{CONCLUSIONS}
\label{sec: con}

The ever-growing increase in the availability of data captured by different sensors coupled with advances in methodological approaches and computational tools makes it desirable to fuse the considerably heterogeneous complementary datasets to increase the efficacy and efficiency of the remotely sensed data processing approaches with respect to the problem at hand. 

The field of multisensor and multitemporal data fusion for remotely sensed imagery is enormously broad which makes it challenging to treat it comprehensively in one literature review. This article focuses particularly on advances in multisource and multitemporal data fusion approaches with respect to different research communities since the methods for the fusion of different modalities have expanded along different paths with respect to each research community. In this context, several vibrant fusion topics, including pansharpening and resolution enhancement, point cloud data fusion, hyperspectral and LiDAR data fusion, multitemporal data fusion, as well as big data and social media were detailed and their corresponding challenges and possible future research directions were outlined and discussed.

As demonstrated through the challenges and possible future research of each section, although the field of remote sensing data fusion is mature, there are
still many doors left open for further investigation, both from the theoretical and application perspectives. We hope
that this review opens up new possibilities for readers to further investigate the remaining challenges to developing sophisticated fusion approaches suitable for the applications at hand.

\section{ACKNOWLEDGMENT}

This research was partially supported by the "High Potential Program" of Helmholtz-Zentrum Dresden-Rossendorf, partially supported by Natural Science Foundation of China under contract 71331005, by State Key Research and Development Program of China under contract 2016YFE0100300. In addition, the authors would like to thank the National Center for Airborne Laser Mapping (NCALM) at the University of Houston for providing the CASI Houston dataset, and the IEEE GRSS Image Analysis and Data Fusion Technical Committee for organizing the 2013 Data Fusion Contest. The authors would like to thanks Dr. Claas Grohnfeldt for processing J-SparseFI-HM. 

\bibliographystyle{IEEEtran}
\bibliography{refsfusion,RSSN,NY,RefMDF}

\end{document}